\documentclass{article}

\usepackage[main,final]{neurips_2026}
\usepackage[utf8]{inputenc}
\usepackage[T1]{fontenc}
\usepackage{url}
\usepackage{graphicx}
\usepackage{booktabs}
\usepackage{amsfonts}
\usepackage{amssymb}
\usepackage{nicefrac}
\usepackage{microtype}
\usepackage{xcolor}
\usepackage{amsmath}
\usepackage{natbib}
\usepackage{listings}
\usepackage{array}
\usepackage{tabularx}
\usepackage{float}
\usepackage{placeins}
\usepackage{capt-of}
\usepackage{enumitem}
\usepackage{hyperref}

\hypersetup{
  colorlinks=true,
  pdftitle={Memory Inception: Latent-Space KV Cache Manipulation for Steering LLMs},
  pdfauthor={Andy Zeyi Liu, Michael Zhang, Ilana Greenberg, Adam Alnasser, Lucas Baker, John Sous},
  linkcolor=blue!50!black,
  citecolor=blue!50!black,
  urlcolor=blue!50!black
}

\newcommand{\R}{\mathbb{R}}
\newcommand{\softmax}{\mathrm{softmax}}
\newcommand{\Attn}{\mathrm{Attn}}
\newcommand{\Norm}{\mathrm{Norm}}
\newcommand{\RoPE}{\textsc{RoPE}}
\newcommand{\concat}{\mathrm{concat}}
\newcommand{\sigmoid}{\sigma}
\newcommand{\mem}{\mathrm{mem}}
\newcolumntype{P}[1]{>{\raggedright\arraybackslash}p{#1}}
\newcolumntype{Y}{>{\raggedright\arraybackslash}X}
\floatstyle{ruled}
\newfloat{algorithm}{t}{loa}
\floatname{algorithm}{Algorithm}

\usepackage{soul}

\lstdefinestyle{pythonstyle}{
  language=Python,
  basicstyle=\ttfamily\footnotesize,
  breaklines=true,
  frame=single,
  columns=fullflexible,
  showstringspaces=false,
  keywordstyle=\color{blue!60!black},
  commentstyle=\color{black!55},
  stringstyle=\color{teal!60!black}
}
\title{\textbf{Memory Inception: Latent-Space KV Cache Manipulation for Steering LLMs}}

\author{%
Andy Zeyi Liu\thanks{Correspondence to: \texttt{andy.liu@yale.edu} and \texttt{john.sous@yale.edu}.}\\
Yale University
\And
Michael Zhang\\
Princeton University
\And
Ilana Greenberg\\
Yale University
\AND
Adam Alnasser\\
Jump Trading
\And
Lucas Baker\\
Jump Trading
\And
John Sous\footnotemark[1]\\
Yale University
}

\date{}

\begin{document}
\maketitle

\begin{abstract}
Prompting and activation steering are commonly used to guide the behavior and personality of LLMs, yet each has its drawbacks: prompting caches repeatedly and is diluted over long context, while activation steering is weaker and struggles to convey nuance. This work introduces \emph{Memory Inception} (MI), which reformulates steering as a selective KV cache allocation problem and thus steers in latent attention space via text-derived key-value banks applied to specific layers. On personality-steering tasks, its effectiveness is similar to prompting and consistently higher than contrastive addition, while offering greater flexibility in mid-conversation shifts without rewriting transcripts. Additionally, it scores higher than both on a post-shift alignment test with Qwen3. MI outperforms prompting on HARDMath and PHYSICS (10/12 subject$\times$mode cells) while cutting added KV storage by up to 118$\times$ on matched content. For steering that requires persistent application and structure, or when transcript manipulation is expensive, MI offers a powerful yet flexible new method. Our code is available \href{https://github.com/AndyLiuin/Memory_Inception}{here}.
\end{abstract}

\section{Introduction}

The standard method of steering language model behavior, whether for crafting personality, altering tone, or persisting facts and preferences for later use, is prompting: placing instructions verbally in context. Activation steering, often used in alignment research, is another commonly seen method that works by directly updating hidden states through learned directions or contrastive residuals. Prompting is powerful and flexible, but often heavy-handed and exhibits diminishing effects over time. Activation steering, in contrast, is compact but rigid, and cannot easily express multi-part or structured behavioral modifications.

\begin{figure*}[t]
    \centering
    \includegraphics[width=\textwidth]{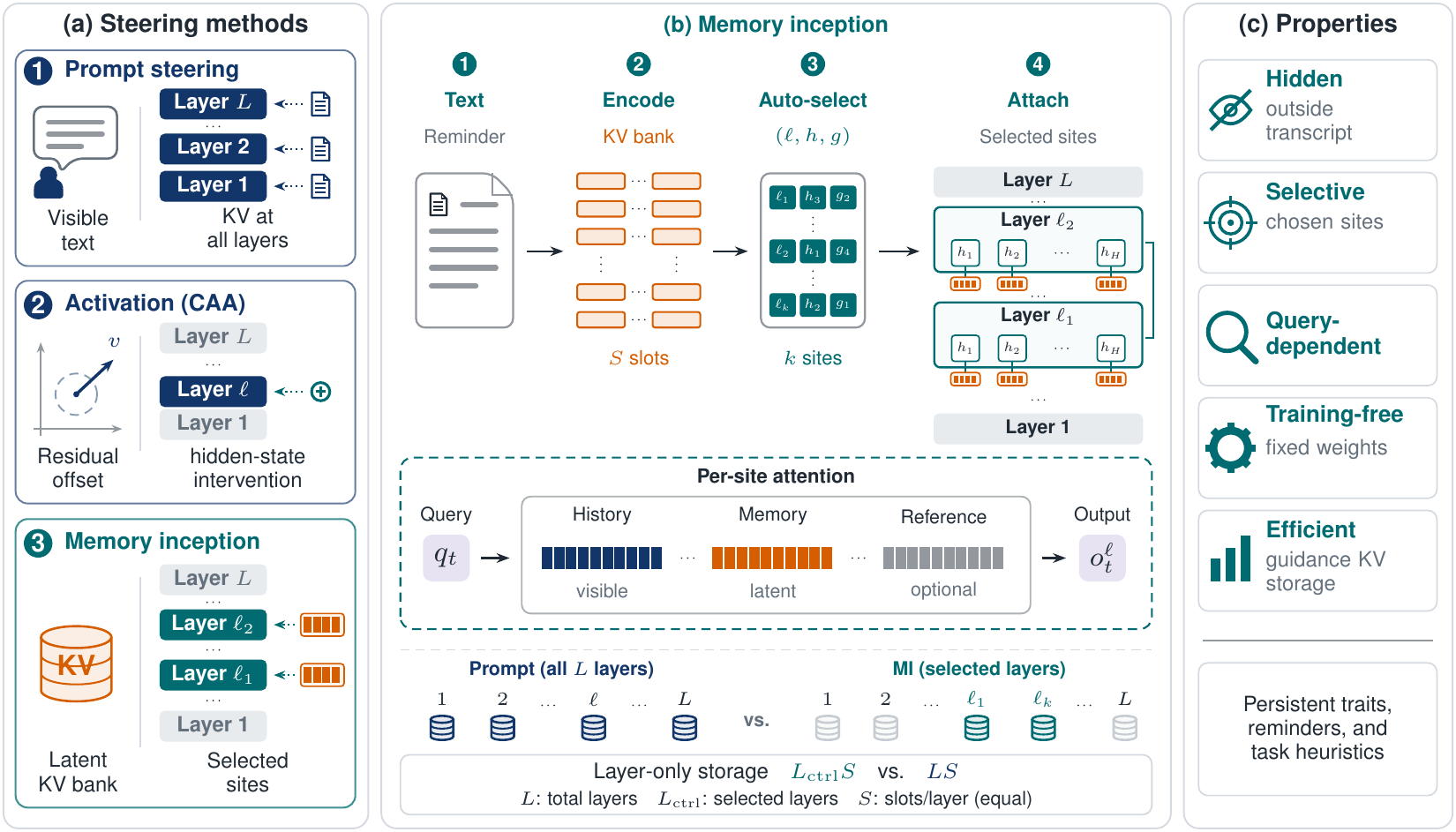}
    \caption{\textbf{Memory inception (MI).}
Reminder text is encoded into latent memory bank slots in the KV cache, then appended to a subset of automatically chosen layers, heads, or key-value groups. MI improves on prompting by keeping the reminder outside the visible transcript. Unlike contrastive activation steering, which acts consistently on specific hidden states, MI can act differently per query based on the attention directed to the reminder slots.}
    \label{fig:memory_bank_attention_steering}
\end{figure*}

In this paper, we study latent attention state as a means of steering, introducing \emph{memory inception} (MI) to balance the power of prompting with the exactitude of activation steering. As illustrated in Figure~\ref{fig:memory_bank_attention_steering}, MI is a training-free steering method that uses latent key-value slots to encode text-based descriptors, facts, and heuristics. The model can then attend to these reminder banks at selected attention sites while keeping steering outside the visible transcript. Rather than impacting all $L$ layers as with prompting, MI targets only the small subset of layers $L_{\rm ctrl}$ chosen to maximize behavioral impact, dramatically shrinking its KV cache footprint.

We evaluate MI across three regimes. Firstly, we evaluate the tradeoff between control and drift on matched personality tasks in which all three methods receive the same control signal. Secondly, we measure the effectiveness of each method on persistent activation and mid-dialogue updates in long conversations. Finally, we test whether the same mechanism can also propagate reasoning techniques in the HARDMath~\citep{fan2024hardmath} and PHYSICS~\citep{feng2025physics} benchmarks. Our main contributions are as follows:

\begin{itemize}[noitemsep,leftmargin=*,topsep=-4pt]
    \item \textbf{Method.} We introduce \emph{memory inception} (MI), a training-free method that encodes reminder text into latent KV banks and makes them available at selected attention sites.

    \item \textbf{Implementation.} We formulate MI as bank-level attention routing. We give a canonical-key \RoPE{} implementation with query-head selection for Llama and KV-group selection for Qwen3.

    \item \textbf{Evaluation.} On the primary personality-steering tasks, MI remains competitive with prompting while outperforming CAA on average drift-adjusted scores. On PHYSICS, MI outperforms prompting in 10/12 subject$\times$mode cells. On HARDMath, MI improves over the plain model but trails prompting overall.

    \item \textbf{Estimated KV storage.} Under content-matched accounting, we estimate reductions of 6--118$\times$ in guidance-derived KV storage relative to visible prompting.
\end{itemize}

\section{Related Work}

The list below covers the methods most directly comparable to MI, with additional related work in Appendix~\ref{app:additional_related_work}.

\textbf{Prompting and latent prompts.}
Prompting is still the most prevalent means of steering language models, from context instructions to chain-of-thought~\citep{brown2020gpt3,wei2022cot,kojima2022zeroshot}. A closely related family of methods learns continuous prefixes without changing model weights \citep{li2021prefixtuning,lester2021prompttuning}. Like prefix-tuning, MI provides latent information that subsequent tokens can attend to. 

\textbf{Activation and attention steering.}
Activation Addition and CAA steer models by adding residual vectors derived from context pairs of positive and negative examples~\citep{turner2023actadd,rimsky2024caa}, but usually this is a few words with one precise meaning. Representation-engineering and function-vector work likewise shows that reusable latent directions can capture high-level behaviors or in-context functions \citep{zou2023repe,todd2024functionvectors}. Related attention-based methods, such as AutoPASTA and Spotlight, steer by allocating more attention toward instruction tokens \citep{zhang2024autopasta,venkateswaran2025spotlight}. In contrast, MI adds text-derived KV slots for attention to read. \citet{belitsky2025kvcache} steer frozen models using contrastive KV vectors from prompts with and without reasoning traces. They shift existing cache entries once after prefill at a designed prompt position across layers.

\section{Memory Inception: Formalism \& Implementation}
\label{sec:method}

Memory inception (MI) is a training-free steering method that encodes reminder content into latent KV banks and attaches those banks only at a small set of selected layers and attention units (we use unit because under different attention mechanism this has different meanings), where the model attends to them alongside ordinary 
prompt tokens. In this section we formalize the procedure. 

\subsection{Memory Inception in Latent Attention Space}
\label{sec:latent-attention}

In this section we introduce the method, starting with the basics: what is exposed to each selected attention head, where additional banks are attached, and how routing changes when more than one memory bank is present.

\paragraph{Attention.}
Consider one attention head in a decoder-only transformer \citep{vaswani2017attention}. At query position $t$, let $T_t$ be the number of available visible-context tokens, including the prompt and generated history. Let the live query be $q_t\in\R^{d_h}$, where $d_h$ is the head dimension, and let the context keys and values be $K^x_{\le t}=[k^x_1,\ldots,k^x_{T_t}]^\top,
V^x_{\le t}=[v^x_1,\ldots,v^x_{T_t}]^\top$. The standard attention output is
\begin{equation}
\Attn(q_t,K^x_{\le t},V^x_{\le t})
=
\softmax\!\left(\frac{q_t {K^x_{\le t}}^\top}{\sqrt{d_h}}\right)V^x_{\le t}.
\label{eq:baseline-attn}
\end{equation}
MI keeps this query-dependent attention computation, but expands the candidate context at selected heads. 

\paragraph{KV Cache Memory Bank.}
A memory bank is a small collection of latent KV slots. Bank $b$ contains $M_b$ slots,
\[
\mathcal{B}^{(b)}=\{(k^{(b)}_m,v^{(b)}_m)\}_{m=1}^{M_b},
\quad
K^{(b)}=[k^{(b)}_1,\ldots,k^{(b)}_{M_b}]^\top,
\quad
V^{(b)}=[v^{(b)}_1,\ldots,v^{(b)}_{M_b}]^\top .
\]
As a reference operation, concatenate the prompt and memory slots:
\begin{equation}
K_t^\star=\concat(K^x_{\le t},K^{(1)},\ldots,K^{(B)}),
\quad
V_t^\star=\concat(V^x_{\le t},V^{(1)},\ldots,V^{(B)}).
\label{eq:augmented-cache}
\end{equation}
Combining the prompt and reminder slots in standard attention gives
$o_t^{\mathrm{concat}}=\Attn(q_t,K_t^\star,V_t^\star)$. MI mixes the prompt and reminder outputs using the rule below. We keep the prompt cache unchanged and store the reminder cache as a separate bank. Only selected layers and attention heads are exposed with this bank. 

\paragraph{Automated Layer and Head Selection.}
Selection is handled automatically based on query-target alignment. Given candidate layers $\mathcal{L}_{\mathrm{cand}}$ and candidate attention units $\mathcal{U}_{\ell}$ within each layer, reference-based calibration scores units by how strongly their queries align with the target bank relative to the reference bank:
\begin{equation}
a_{\ell,u}
=
\max_j \frac{q_{\ell,u}^{\top}k^+_{\ell,u,j}}{\sqrt{d_h}}
-
\max_j \frac{q_{\ell,u}^{\top}k^-_{\ell,u,j}}{\sqrt{d_h}}.
\label{eq:alignment-margin}
\end{equation}
The selector keeps the top $k$ heads or KV groups in each layer, combines their scores to rank the layers, and keeps the top $m$ layers. It saves these choices and keeps them fixed during inference. MI then allows only the selected heads to attend to the reminder bank. All models we test on in this work use grouped-query attention (GQA). The selector chooses individual query heads for Meta-Llama-3.1-8B-Instruct and KV groups for Qwen3-30B-A3B, exposing the bank to all query heads in each selected group. Appendix~\ref{app:selector} gives the detailed algorithms.

\paragraph{Bank-level mixture.}
We express attention at each selected head in two stages: attention within
each bank, followed by a weighted mixture of the bank outputs.
Let $b=0$ denote the prompt bank, with $M_0=T_t$ slots, and
$b=1,\ldots,B$ denote the memory banks. For each bank, define the slot
scores, within-bank attention weights, and output:
\[
s^{(b)}_{t,m}=\frac{\langle q_t,k^{(b)}_m\rangle}{\sqrt{d_h}},
\quad
\alpha^{(b)}_{t,m}=\frac{\exp(s^{(b)}_{t,m})}{\sum_r\exp(s^{(b)}_{t,r})},
\quad
u_t^{(b)}=\sum_m \alpha^{(b)}_{t,m}v^{(b)}_m.
\]
Let $c_t^{(b)}\in\R$ be a bias added to every slot score in bank $b$, and define
the bank evidence as
$\widetilde{\beta}_t^{(b)}
=\log\!\left(\sum_{m=1}^{M_b}\exp(s^{(b)}_{t,m})\right)+c_t^{(b)}$.
Concatenated attention then decomposes exactly as
\begin{equation}
\widetilde{\pi}_t
=
\softmax([\widetilde{\beta}_t^{(0)},\ldots,
         \widetilde{\beta}_t^{(B)}]),
\quad
o_t^{\mathrm{concat},c}
=
\sum_{b=0}^{B}\widetilde{\pi}_t^{(b)}u_t^{(b)}.
\label{eq:bank-mixture}
\end{equation}
Setting all $c_t^{(b)}$ to zero recovers ordinary concatenated attention. To normalize for bank size, we subtract $\log M_b$ from each bank's evidence:
\begin{equation}
\beta_t^{(b)}
=
\widetilde{\beta}_t^{(b)}-\log M_b
=
\log\!\left(\frac{1}{M_b}\sum_{m=1}^{M_b}
\exp(s^{(b)}_{t,m})\right)+c_t^{(b)}.
\label{eq:bank-evidence}
\end{equation}
The resulting bank probabilities and head output are
\begin{equation}
\pi_t
=
\softmax([\beta_t^{(0)},\ldots,\beta_t^{(B)}]),
\quad
o_t^\star
=
\sum_{b=0}^{B}\pi_t^{(b)}u_t^{(b)}.
\label{eq:normalized-bank-mixture}
\end{equation}
When banks differ in size, this normalization method changes their relative weights in a way equivalent to concatenated attention with a bias per slot of $c_t^{(b)}-\log M_b$. Section~\ref{sec:bank-construction} specifies these terms for different objectives. 

\subsection{Constructing Memory Banks}
\label{sec:bank-construction}

Section~\ref{sec:latent-attention} describes how MI works and details how attention KV cache is modified during inference. We now describe how reminder KV caches are constructed from the corresponding text. 

\paragraph{Text-Derived Memory Banks.}
Banks can be built from various sources. To do so, first, the source text is placed inside task-specific templates. We run each variant through the frozen base model and retain only the hidden states at descriptor-token positions. The state $h^{(b)}_{\ell,m}$ is the residual stream entering decoder layer $\ell$, before that layer's input RMSNorm. For each selected layer $\ell$ and selected key-value unit $u$, we use the layer's input RMSNorm $\Norm_\ell$ and key and value projections $W^{(\ell)}_{K,u}$ and $W^{(\ell)}_{V,u}$. The key normalization $\Norm^K_\ell$ is the layer's \texttt{k\_norm} for Qwen3 and the identity for Llama-3.1. The stored keys and values are
\begin{equation}
k^{(b)}_{\ell,u,m}=\Norm^K_\ell\!\left(W^{(\ell)}_{K,u}\Norm_\ell(h^{(b)}_{\ell,m})\right),
\qquad
v^{(b)}_{\ell,u,m}=W^{(\ell)}_{V,u}\Norm_\ell(h^{(b)}_{\ell,m}).
\label{eq:text-bank}
\end{equation}
The keys are stored after any key normalization and before \RoPE{}. Big Five banks retain descriptor states from wrapped variants. PHYSICS and HARDMath also include an unwrapped descriptor variant. Wrapper tokens provide context during encoding but do not become bank slots. Appendix~\ref{app:bank_construction_details} gives the template strings, and Appendix~\ref{app:wrapper_ablation} reports the wrapper ablation. For reasoning tasks, a teacher model first generates the heuristic text, which is then processed similar to above. 

\paragraph{Canonical Pre-\RoPE{} Key Storage.}
Llama-style and Qwen3-style models use rotary position embeddings (RoPE)\citep{su2021roformer,grattafiori2024llama3}. If we treat this part of additional KV cache as normal, this will have position dependence relative to other text and causes complication. Thus we store the memory keys before RoPE. Let $R_t$ be the rotary operator at query position $t$. $q_t=R_t\bar q_t$ and prompt keys have the form $k^x_j=R_j\bar k^x_j$. For a canonical key $\tilde k_m$ and optional relative phase $\delta_m$, the memory score is
\begin{equation}
s^{\mem}_{t,m}
=
\frac{\langle R_t^{-1}q_t,\;R_{\delta_m}\tilde k_m\rangle}{\sqrt{d_h}}
=
\frac{\langle \bar q_t,\;R_{\delta_m}\tilde k_m\rangle}{\sqrt{d_h}} .
\label{eq:rope-memory-score}
\end{equation}
In this way we cancel the query's absolute rotary phase; the default relative phase is $\delta_m=0$. The query and memory representations still reflect their input contexts, so the model’s response remains context-dependent. Bank KV tensors are stored per KV head and each selected query head reads its group's KV head.

\paragraph{Target, Reference, and Auxiliary Banks.}
Many steering tasks benefit from multiple banks with different roles. We use target banks for desired behavior, optional reference banks for opposite behavior, and auxiliary banks for additional facts or heuristics. PHYSICS and HARDMath use target and auxiliary planning banks without a reference bank. Dialogue shifts use explicit opposite-style reference banks. The following contrastive formulation applies when a reference bank is present.
\begin{equation}
\Delta_t=
\log\!\left(\frac{1}{M^+}\sum_m e^{s^+_{t,m}}\right)
-
\log\!\left(\frac{1}{M^-}\sum_n e^{s^-_{t,n}}\right).
\label{eq:delta-bank}
\end{equation}
We convert that gap into query-adaptive gates
$g_t^\pm=\sigmoid(\pm\gamma\Delta_t)$, where
$\sigmoid(z)=(1+e^{-z})^{-1}$, $\gamma\ge0$, and $g_t^\pm\in(0,1)$.
Let $\rho_\ell\ge0$ be the layer gain and $\lambda_+,\lambda_-\ge0$
the target and reference gain magnitudes. In the full-prompt formulation,
the evidence scores are
\begin{equation}
\begin{aligned}
\beta_t^x = \log &\!\left(\frac{1}{T_t}\sum_j e^{s^x_{t,j}}\right),\\
\beta_t^+ = \log\!\left(\frac{1}{M^+}\sum_m e^{s^+_{t,m}}\right)+\rho_\ell\lambda_+ g_t^+&,\quad\beta_t^- = \log\!\left(\frac{1}{M^-}\sum_n e^{s^-_{t,n}}\right)-\rho_\ell\lambda_- g_t^- .
\end{aligned}
\label{eq:contrastive-betas}
\end{equation}
An auxiliary bank $a$ uses $c_t^{(a)}=\rho_\ell\lambda_a$, with $\lambda_a\ge0$, in Eq.~\eqref{eq:bank-evidence}. Without a reference bank, we neglect its terms and set $g_t^+=1$, this gives $c_t^+=\rho_\ell\lambda_+$. 

\paragraph{Qwen3 vLLM approximation.}
The zero-score vLLM path for Qwen3 reuses the native attention output
$o_t^{\mathrm{native}}$ as the prompt-bank output and assigns that bank
a synthetic evidence score of zero. For each active memory bank $b$,
it computes the output $u_t^{(b)}$ from the side-bank slots and uses
\begin{equation}
\widehat{\beta}_t^{(0)}=0,
\qquad
\widehat{\beta}_t^{(b)}
=
\log\!\left(\frac{1}{M_b}\sum_{m=1}^{M_b}
\exp(s_{t,m}^{(b)})\right)
+\widehat{c}_t^{(b)},
\quad b=1,\ldots,B,
\label{eq:qwen-vllm-evidence}
\end{equation}
where $\widehat{c}_t^{(b)}\in\R$ collects the backend's configured
memory-bank gains. The resulting mixture is
\begin{equation}
\begin{aligned}
\widehat{\pi}_t
&=
\softmax([\widehat{\beta}_t^{(0)},\ldots,
         \widehat{\beta}_t^{(B)}]),\\
\widehat{o}_t
&=
\widehat{\pi}_t^{(0)}o_t^{\mathrm{native}}
+
\sum_{b=1}^{B}\widehat{\pi}_t^{(b)}u_t^{(b)}.
\end{aligned}
\label{eq:qwen-vllm-mixture}
\end{equation}
The mixed head outputs are written back before the attention output
projection. Appendix~\ref{app:vllm} gives the implementation details.

\subsection{KV Cache}

Prompting keeps KV cache for every block layer in the transformer. With $T_{\rm prompt}$ prompt tokens and $L$ layers, this requires storage proportional to $LT_{\rm prompt}$. MI stores $S_{\rm bank}$ memory slots at $L_{\rm ctrl}$ selected layers and this requires storage proportional to $L_{\rm ctrl}S_{\rm bank}$. The prompt-to-bank storage ratio is then
\begin{equation}
\mathrm{KV\ ratio}
=
\frac{L T_{\mathrm{prompt}}}{L_{\mathrm{ctrl}}S_{\mathrm{bank}}}.
\label{eq:kv-ratio}
\end{equation}
For Qwen3-30B-A3B, $L=48$ and $L_{\rm ctrl}=5$, giving a ratio of 0.6 when the bank has as many slots as the native prompt. When we actually calculate resources in later evaluation, we account for all slots across the attached banks. During our actual evaluation, if a MI bank is shared, it is stored only once. Appendix~\ref{app:measured_efficiency} reports measured tensor sizes and timing results separately from these storage estimates.

\section{Experimental Design}

\paragraph{Models and Baselines.}
Our main experiments use Meta-Llama-3.1-8B-Instruct~\citep{meta2024llama31modelcard} and
Qwen3-30B-A3B~\citep{qwen2025qwen3modelcard}, with additional personality steering evaluation on gpt-oss-120b~\citep{openai2025gptoss} in Appendix~\ref{app:gptoss}. As baselines, we compare the plain model without special guidance ("Plain"), ordinary prompting with guidance visible in the transcript ("Prompt"), and activation steering with CAA. The reasoning experiments leave out CAA due to its unsuitability for long or complex steering instructions. 

The automated selector chooses individual query heads for Llama and KV groups for Qwen3. In Qwen3, selection may target different KV groups in each affected layer. We use vLLM~\citep{kwon2023pagedattention} for batched generation on Qwen3, with standard inference for Plain and Prompt modes and method-specific interventions for CAA and MI. 

\paragraph{Tasks and metrics.}
We evaluate models in four settings:
\begin{itemize}[noitemsep,leftmargin=*,topsep=2pt]

  \item \textbf{Big Five self-evaluation.} The model answers public-domain IPIP-50-style items~\citep{goldberg2006ipip}, giving a Likert-scale score from 1-5 for each item. We measure target-trait shift compared to the plain model, $\Delta_{\mathrm{target}}$, as well as mean absolute drift in other traits $D_{\mathrm{non\text{-}target}}$. The \textit{drift-adjusted score} subtracts this drift from the target shift $\mathrm{DAS}=\Delta_{\mathrm{target}}-D_{\mathrm{non\text{-}target}}$.

  \item \textbf{Big Five generation.} In this case we steer the same traits in open-ended text and use GPT-4o-mini to judge the scores of the target and non-target traits, along with text coherence. The main metric here is the \textit{generation-adjusted score} $\mathrm{GAS}=\Delta_{\mathrm{target}}-D_{\mathrm{non\text{-}target}}$, where $\Delta_{\rm target}$ is the target-score shift compared to the plain model and $D_{\rm non-target}$ measures drift in other traits. Contrastive margin and coherence are auxiliary diagnostics 
  (Appendix~\ref{app:bigfive_generation_details}).

  \item \textbf{Dialogue shift.} In this evaluation we test whether the controller text can change the model's behaviour midway through a conversation. The benchmark contains 204 branching cases: 34 scenarios, each evaluated at 5, 16 and 24 turns with two branches. Scenarios cover support, operations, skill building and planning. We report post-shift alignment, $s_{\rm target}-s_{\rm old}$, which compares alignment with the new and previous targets, as well as judged response quality. We similarly use GPT-4o-mini as judge. Further construction details are in Appendix~\ref{app:benchmarks}.

  \item \textbf{HARDMath and PHYSICS.} These benchmarks cover applied mathematics~\citep{fan2024hardmath} and university-level physics~\citep{feng2025physics}. MI receives task heuristics as memory banks. We report normalized benchmark scores under each task's grading pipeline. GPT 5.5 is used as judge. (Further details are described in Appendices~\ref{app:hardmath_benchmark} and~\ref{app:physics_benchmark}).
\end{itemize}

\paragraph{Memory-bank construction by task.}
We construct the banks before evaluation, banks vary across different tasks. For Big Five tasks, the bank encodes a description of the target trait, e.g. "extraverted". Questionnaire items and generation prompts are excluded from bank construction. For dialogue tasks, the bank encodes the requested target style, optionally with the previous style as a reference. For HARDMath and PHYSICS, we ask GPT-5.5-thinking to extract reasoning heuristics from five construction examples per subject. We only keep natural-language guidance on the reasoning process. We exclude these construction examples from evaluation in both benchmarks. Appendix~\ref{app:memory_bank_examples} gives representative examples.

\section{Main Results}

\begin{table*}[!t]
\centering
\small
\setlength{\tabcolsep}{3pt}
\renewcommand{\arraystretch}{1.12}
\caption{\textbf{Main results.} Method columns differ by block. Higher scores are better for all reported metrics. Bold marks the highest steering score, reasoning score.}
\label{tab:core_tradeoff}
\begin{tabularx}{\textwidth}{@{}P{0.20\textwidth}P{0.18\textwidth}P{0.19\textwidth}*{3}{>{\raggedleft\arraybackslash}X}@{}}
\toprule
Task & Model & Setting / metric & \multicolumn{3}{c}{Method} \\
\midrule
\multicolumn{3}{l}{\textbf{Steering}} & Prompt & CAA & Ours \\
\cmidrule(lr){4-6}
Big Five self-eval & \mbox{Llama3.1-8B} & $\mathrm{DAS}$ & 0.645 & 0.185 & \textbf{0.775} \\
& \mbox{Qwen3-30B-A3B} & $\mathrm{DAS}$ & \textbf{0.355} & 0.115 & 0.270 \\
\addlinespace[2pt]
Big Five generation & \mbox{Llama3.1-8B} & $\mathrm{GAS}$ & \textbf{-0.524} & -0.874 & -0.602 \\
& \mbox{Qwen3-30B-A3B} & $\mathrm{GAS}$ & 0.260 & -0.926 & \textbf{0.440} \\
\midrule
\multicolumn{3}{l}{\textbf{Updateable guidance} (Align. / Quality)} & Prompt & CAA & Ours \\
\cmidrule(lr){4-6}
Dialogue shift & \mbox{Qwen3-30B-A3B} & Overall & \mbox{0.438 / 8.54} & \mbox{0.526 / 8.44} & \mbox{\textbf{0.816} / 7.75} \\
& \mbox{Qwen3-30B-A3B} & 24 turns & \mbox{0.344 / 8.54} & \mbox{0.497 / 8.37} & \mbox{\textbf{0.651} / 7.80} \\
\addlinespace[2pt]
& \mbox{Llama3.1-8B} & Overall & \mbox{1.178 / 7.58} & \mbox{\textbf{1.554} / 7.34} & \mbox{1.516 / 7.00} \\
& \mbox{Llama3.1-8B} & 24 turns & \mbox{0.964 / 7.76} & \mbox{\textbf{1.380} / 7.43} & \mbox{1.344 / 7.03} \\
\midrule
\multicolumn{3}{l}{\textbf{Structured reasoning}} & Plain & Prompt & Ours \\
\cmidrule(lr){4-6}
PHYSICS & \mbox{Qwen3-30B-A3B} & no-CoT avg. (\%) & 67.5 & 69.6 & \textbf{70.8} \\
& \mbox{Qwen3-30B-A3B} & CoT avg. (\%) & 80.3 & 81.2 & \textbf{82.9} \\
& \mbox{Qwen3-30B-A3B} & Overall avg. (\%) & 73.9 & 75.4 & \textbf{76.9} \\
& \mbox{Qwen3-30B-A3B} & Win count (12 cells) & 1 & 1 & \textbf{10} \\
\addlinespace[2pt]
HARDMath & \mbox{Qwen3-30B-A3B} & CoT overall (\%) & 61.0 & \textbf{67.3} & 63.8 \\
\bottomrule
\end{tabularx}
\end{table*}

\begin{figure*}[t]
\centering
\includegraphics[width=\textwidth]{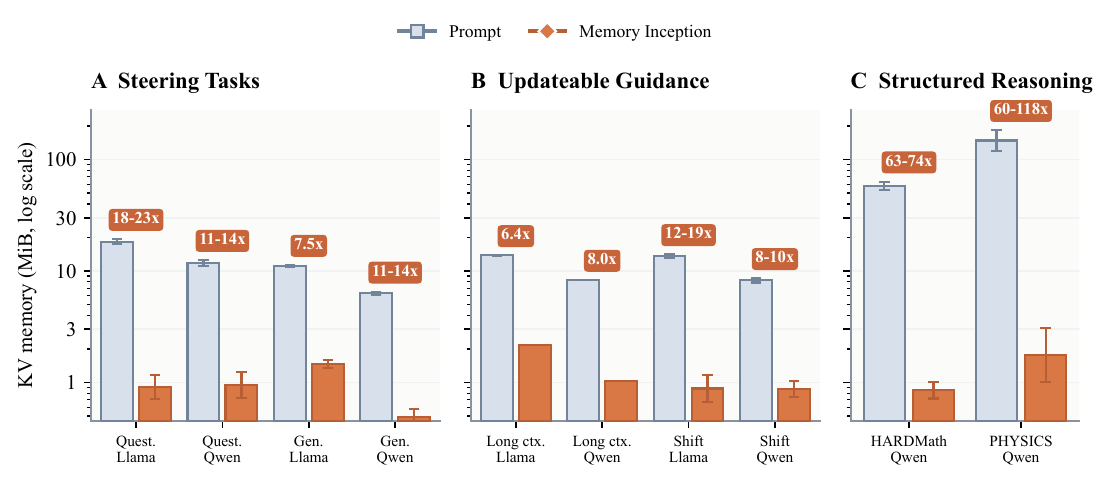}
\caption{\textbf{Estimated KV cache storage.} Grey bars show the KV cache storage consumed by prompting. In comparison, orange bars show MI's KV cache inferred from content-matched accounting. We also report the prompt-to-bank storage ratios. The vertical axis uses a log scale in MiB.}
\label{fig:cache_cost}
\end{figure*}

In Table~\ref{tab:core_tradeoff} we summarize task performance and in Figure~\ref{fig:cache_cost} we report the estimated KV-cache storage. 

\paragraph{Personality steering.} For the Big Five evaluation, MI outperforms CAA on DAS and GAS for both Llama and Qwen3. These metrics measure target-trait movement after subtracting drift in the other traits. The comparison with prompting depends on the model and task. MI leads on Llama self-evaluation and Qwen3 generation, while Prompt leads on Qwen3 self-evaluation and Llama generation. MI therefore provides better drift-adjusted steering than CAA in these experiments, with comparable performance against Prompt (In Appendices~\ref{app:bigfive_self_eval_details} and~\ref{app:bigfive_generation_details} we provide the detailed per-trait breakdown).

\paragraph{Guidance updating during dialogue. }For the dialogue results, MI achieves the highest post-shift alignment on Qwen3, both overall and at 24 turns, compared to Prompt and CAA. On Llama, MI's alignment is comparable to CAA and both are much better than Prompt. Prompting is susceptible to context contamination, where previous context has an unwanted impact on downstream output, making post-shift control harder and less consistent. In this type of scenario, MI's lack of contamination provides a clear advantage.

\paragraph{Reasoning with heuristic banks. }The reasoning experiments compare heuristic banks supplied as explicit text at the start of the prompt versus as a persistent latent bank. The guidance combines subject heuristics, global planning, and a completion checklist. We compare Plain, Prompt and MI since CAA does not have a comparable way to inject long heuristic banks.

MI achieves the highest overall score on PHYSICS and wins 10 of 12 subject cells, with and without CoT. On HARDMath, MI improves over Plain overall but slightly trails Prompt. We surmise that the improvement owes mainly to MI's consistency in carrying heuristic banks throughout inference, acting as a persistent reminder, while Prompt has the desired impact initially but fades with length.

\paragraph{KV-Cache Savings and Measured Efficiency.}
Figure~\ref{fig:cache_cost} reports estimated reductions in added KV storage of 6.4$\times$ in the long-context Llama
case, 60--118$\times$ on PHYSICS, and 63--74$\times$ on HARDMath, matching banks according to content. For Qwen3, example measured tensor payloads are 1.904\,MiB for a PHYSICS subject bank and
8.906\,MiB for a dialogue target-plus-reference bank. MI also reduces the throughput impact of guidance relative to Prompt: in a 234-example HARDMath timing run on a single H200, MI generates 51.5 output tokens/s, compared with 51.4 for Plain and 46.4 for Prompt. Due to higher output length, MI takes 16.43\,h
overall, compared with 13.27\,h for Prompt. Appendix~\ref{app:measured_efficiency} gives more detail on tensor counts and timing protocol.

\paragraph{Additional results.}
On the additional MoE-based gpt-oss-120b, MI achieves comparable performance to Prompt on both the Big Five questionnaire and generation tasks. MI has lower KV-cache storage. Detailed results are in Appendix~\ref{app:gptoss}.

\begin{figure*}[t]
\centering
\includegraphics[width=\textwidth]{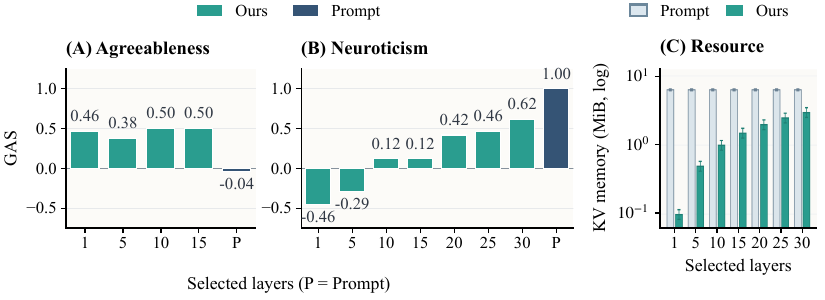}
\caption{\textbf{Qwen3 layer-selection ablation.} Panels A and B report GAS for an easier trait (agreeableness) and a harder trait (neuroticism) as the number of selected layers increases. Panel C shows the corresponding budget-equivalent KV footprint relative to prompting.}
\label{fig:layer_ablation}
\end{figure*}
\begin{figure*}[t]
\centering
\includegraphics[width=0.8\textwidth]{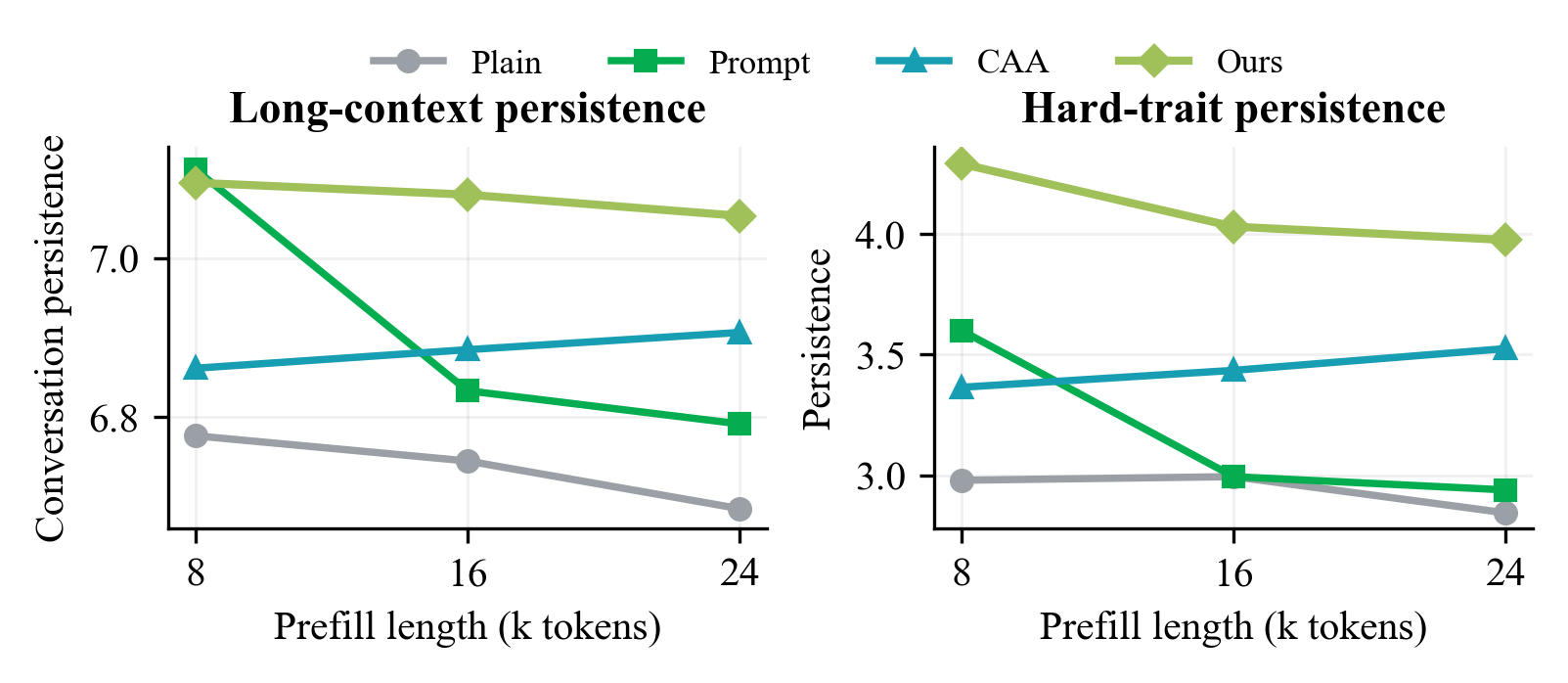}
\caption{\textbf{Qwen3 long-context persistence under opposite-style prefills.} Left: average persistence across all six target traits. Right: the harder \texttt{dismissive}+\texttt{anxious} average. The full Qwen3 trait$\times$prefill breakdown is in Appendix Table~\ref{tab:long_context_trait_budget_details}, and the corresponding Llama breakdown is in Appendix Table~\ref{tab:llama_long_context_trait_budget_details}.}
\label{fig:persistent_guidance}
\end{figure*}

\section{Analysis and Ablations}

\paragraph{Setup and scope.} In this section we perform two ablation studies on Qwen3-30B-A3B. The first is a \textbf{layer-selection} ablation (Figure~\ref{fig:layer_ablation}), varying the number of selected layers while holding bank content fixed. We investigate how much latent coverage MI needs for traits of different difficulties. The second is a \textbf{long-context persistence} ablation (Figure~\ref{fig:persistent_guidance}): we append a long opposite-style prefill before the actual inference, testing how well each method's signal survives a noisy prefill.

\paragraph{Layer Ablation.} In this case we test two personality traits that present different levels of steering difficulty. In Qwen3's case, \textit{agreeableness} aligns well with its default behavior and only requires a small nudge, while \textit{neuroticism} runs contrary to its normal personality and is harder to induce. For the easier trait, \textit{agreeableness}, a single-layer intervention achieves the desired effect, and scores stabilize when MI is extended to 10 layers. For \textit{neuroticism}, fewer than 10 layers leave the score negative, and meaningful control requires attaching the bank to more layers. Even at the widest settings, however, MI provides significant savings in memory relative to direct prompting.

\paragraph{Long-Context Persistence.}
We also tested how increasing the length of opposite-style prefill affects steering, using the same dialogue cases. On Qwen3, prompt-based persistence declines as the prefill grows, while MI's average across six traits remains comparatively stable (Figure~\ref{fig:persistent_guidance}, left). The difference is larger for the harder \textit{dismissive} and \textit{anxious} traits (right). CAA is more stable than Prompt, but MI maintains the highest persistence on these two traits across the tested prefill lengths. MI's dismissive persistence also declines with longer prefills, so the average stability does not hold for every trait. Appendix~\ref{app:benchmarks} describes the setup, and Appendix Tables~\ref{tab:long_context_trait_budget_details} and~\ref{tab:llama_long_context_trait_budget_details} give the full results.

\paragraph{General Capability Check.}
In addition, we perform general a capability check on GSM8K and MMLU after steering towards various personas. Appendix~\ref{app:capability} confirms that MI produces no meaningful capability degradation relative to the plain model, and less degradation than visible prompting across trait-benchmark pairs where drops occur.

\FloatBarrier
\section{Conclusion}

As a training-free method for LLM steering that works by appending KV banks derived from guidance text to selected attention layers, memory inception (MI) provides a middle ground between visible prompting and activation steering that largely combines the power and flexibility of the former with the efficiency of the latter. MI improves personality steering on Big Five evaluation relative to CAA, with effects comparable to direct prompting. By detaching and replacing old cache adjustments, it also supports flexible mid-conversation behavioral shifts without changing the transcript, while maintaining the ability to provide consistent steering throughout longer interactions. MI is highly effective in long and reasoning-intensive tasks such as those in PHYSICS and HARDMath, where it outperforms direct prompting. Across all settings, MI reduces KV cache storage impact by a factor of up to 118$\times$ compared to visible prompting, making it an attractive choice for persistent, structured, or complex guidance that should be applied uniformly without polluting context.

\paragraph{Limitations.}
We note three limitations: scoring methods, task variability, and model variability. Firstly, our task families use different judges and scoring protocols, with LLM-scored Likert shifts for personality tasks and rubric-based partial credit on reasoning tasks. We report these separately as they are not directly comparable. Secondly, bank quality depends on both task and phrasing: as with ordinary prompting, guidance that is too narrow, too broad, irrelevant, or otherwise flawed may fail to produce the desired result. Thirdly, different models may show different responses or sensitivity levels. In our tests, Qwen3 shows the clearest gains from persistent guidance, while Llama exhibits sharper tradeoffs between quality and steerability.

We also caveat that our results do not rigorously establish end-to-end latency or VRAM savings. Finally, relative to visible prompting, MI could be misused to induce covert behavioral shifts or hidden prompt injections, making disclosure and auditability of steering methods an important factor in real-world deployment.

\paragraph{Responsible release.}
Since latent steering can be used to hide prompt injections that would ordinarily be visible, we provide only research code, evaluation, and benchmark assets, which may require further adaptation before deployment. For real use, we strongly recommend that latent steering methods including MI be disclosed to providers and users, with access control, audit logging, and human review prior to release.

\label{end:main_text}

\begin{ack}
This research was supported in part by the Yale Office of the Provost. We thank AMD for providing GPU computing resources used in this work. Experiments were carried out on the Yale Bouchet Cluster and Runpod. A.~Alnasser and L.~Baker were supported by Jump Trading.
The authors declare no competing financial interests beyond the affiliations and support disclosed above.
\end{ack}

\bibliographystyle{plainnat}
\bibliography{ref}

\clearpage
\appendix
\raggedbottom

\section*{Appendix contents}
\begin{itemize}[leftmargin=1.5em,itemsep=0.2em,topsep=0.25em]
    \item \hyperref[app:additional_related_work]{Additional related work}
    \item \hyperref[app:model_cards]{Models, implementation, and efficiency}
    \item \hyperref[app:selector]{Automated attention head/layer selection}
    \item \hyperref[app:benchmarks]{Datasets and evaluation}
    \item \hyperref[app:memory_bank_examples]{Memory-bank construction and examples}
    \item \hyperref[app:results_breakdown]{Results breakdown by task}
    \item \hyperref[app:technical_derivations]{Technical derivations}
\end{itemize}

\section{Additional related work}
\label{app:additional_related_work}

Beyond the prompting and activation-steering baselines in the main text, MI is also close to a broader family of inference-time memory mechanisms. Retrievel-augmented generation, memorizing transformers and RETRO augment the model with external memory at inference time~\citep{lewis2020rag,khandelwal2020knnlm,wu2022memorizing,borgeaud2022retro}. These methods mainly target factual recall or LLM quality, whereas MI uses compact additional banks to carry behavior or reasoning heuristics. 

There are several other works adjacent to activation steering, such as Activation Scaling, Improved Representation Steering, and conditional activation steering. They refine how steering directions are normalized, composed, and conditionally activated \citep{stoehr2024activationscaling,wu2025reps,lee2025cast}. These methods support the general claim that useful control can be achieved without retraining, but they still operate through intervention on the activation. 

Recent attention-path methods are close in spirit. Spotlight dynamically increases attention to designated instructions, while SEKA and PRISM-$\Delta$ steer key representations or attention subspaces to highlight visible prompt spans \citep{venkateswaran2025spotlight,li2026seka,ge2026prism}. KV Cache Steering intervenes directly in the cache to induce controlled behavior in otherwise frozen models \citep{belitsky2025kvcache}. MI differs from these approahces in its interface: insted of reweighting visible prompt tokens, it stores information as latent side-banks that can be updated independently of the visible transcript.

\section{Models, implementation, and efficiency}
\label{app:model_cards}

\paragraph{Models.}
Table~\ref{tab:model_cards} lists the three backbones, using their public
model cards and technical reports~\citep{meta2024llama31modelcard,
grattafiori2024llama3,qwen2025qwen3modelcard,qwen2025qwen3technicalreport,
openai2025gptoss}. The layer count, number of KV heads, and head dimension
determine how much memory a prompt token or a bank slot occupies
(Appendix~\ref{app:measured_efficiency}).

\begin{table*}[!htbp]
\centering
\small
\setlength{\tabcolsep}{3pt}
\caption{Backbone models. Active parameters are those used per token.}
\label{tab:model_cards}
\begin{tabularx}{\linewidth}{@{}P{0.21\linewidth}ccccccYY@{}}
\toprule
Model & Total & Active & \shortstack{Experts\\(active)} & Layers
  & \shortstack{Q / KV\\heads} & \shortstack{Head\\dim} & Context & License \\
\midrule
Llama-3.1-8B-Instruct & 8B & 8B & dense & 32 & 32 / 8 & 128
  & 128k & Llama 3.1 Community \\
Qwen3-30B-A3B & 30.5B & 3.3B & 128 (8) & 48 & 32 / 4 & 128
  & 32k (131k w/ YaRN) & Apache 2.0 \\
gpt-oss-120b & 116.8B & 5.1B & 128 (4) & 36 & 64 / 8 & 64
  & 128k & Apache 2.0 \\  
\bottomrule
\end{tabularx}
\end{table*}

\paragraph{Hardware.}
Experiments ran on NVIDIA H100, H200, B200, and RTX Pro 6000 Blackwell
GPUs, depending on cluster availability. Model, bank, and method settings
are fixed within each sweep, but different parts of a sweep may have run
on different GPU types. All timing results
(Appendix~\ref{app:measured_efficiency}) come from a single H200.

\begin{figure*}[!htbp]
    \centering
    \includegraphics[width=\textwidth]{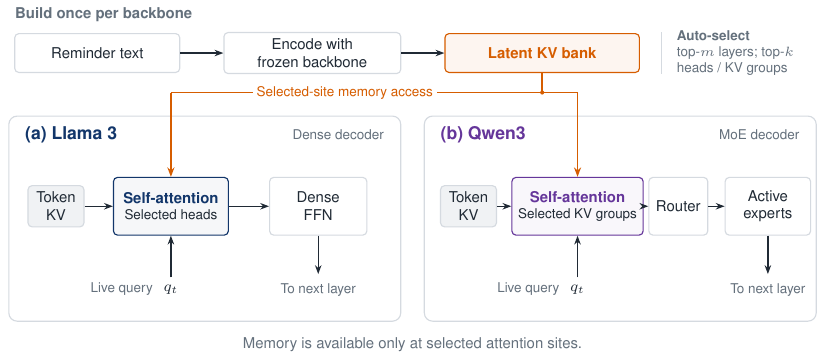}
    \caption{How MI attaches a memory bank in each architecture. Each
    frozen model encodes the reminder text into bank keys and values. The
    selector (Appendix~\ref{app:selector}) chooses which layers and
    attention units read from the bank: individual query heads in Llama,
    and KV groups in Qwen3, where steering applies to every query head in
    the group. The steered attention output then passes through Llama's
    dense FFN or Qwen3's MoE router and experts as usual. Normalization
    and residual connections are omitted. The diagram is conceptual: the
    Qwen3 vLLM implementation stores the bank separately and does not
    modify the native paged KV cache.}
    \label{fig:architecture_attention_steering}
\end{figure*}

\subsection{vLLM inference implementation}
\label{app:vllm}

We run Qwen3 in vLLM for batched inference. Plain and Prompt use the
unmodified model. CAA adds a per-layer vector to the attention output
after the output projection, before the residual addition.

MI changes only the layers chosen by the selector. Bank keys and values
are kept in a separate buffer, so the native paged KV cache is unchanged.
At each selected layer, for the query heads in the selected KV groups, MI:
\begin{enumerate}
  \item computes native attention over the context as usual, giving
        $o_t^{\mathrm{native}}$;
  \item scores the bank keys against the normalized, pre-\RoPE{} query
        and computes each bank's output $u_t^{(b)}$ and evidence
        $\widehat{\beta}_t^{(b)}$ (Eq.~\eqref{eq:qwen-vllm-evidence});
  \item assigns the native output a fixed evidence of zero instead of
        computing the context's score, which we call the
        \emph{zero-score approximation}, and mixes the outputs with
        Eq.~\eqref{eq:qwen-vllm-mixture};
  \item writes the mixed output back before the attention output
        projection.
\end{enumerate}

\subsection{gpt-oss-120b implementation}
\label{app:gptoss_impl}

MI on gpt-oss-120b~\citep{openai2025gptoss} follows the Qwen3
implementation, with steering applied to the query heads in each selected
KV group. Three architectural differences require changes:
\begin{itemize}
  \item \textbf{Bank construction.} We keep the K/V projection biases,
        store keys before \RoPE{}, and skip key normalization, which
        gpt-oss does not have.
  \item \textbf{Sliding-window layers.} gpt-oss alternates windowed and
        full attention. At selected windowed layers, the window limits
        only context tokens; bank slots stay visible.
  \item \textbf{Attention sinks.} gpt-oss has learned sink logits. A sink
        behaves like a slot with a learned score and a zero value vector.
\end{itemize}
The expert router and all model weights are unchanged. For gpt-oss we
report only Big Five self-evaluation and generation
(Appendix~\ref{app:gptoss}).

\subsection{Memory and throughput}
\label{app:measured_efficiency}

\paragraph{Bank size.}
A bank slot stores one key and one value for each selected layer and KV
group. For Qwen3, with 5 selected layers, 4 KV groups (all four per
layer), head dimension 128, and bf16 (2 bytes),
\[
\underbrace{5}_{\text{layers}}\times\underbrace{4}_{\text{KV groups}}
\times\underbrace{128}_{\text{head dim}}\times\underbrace{2}_{K,V}
\times\underbrace{2}_{\text{bytes}} = 10{,}240\ \text{bytes (10\,KiB) per slot}.
\]
A prompt token in Qwen3's ordinary KV cache occupies
$48\times4\times128\times2\times2 = 98{,}304$ bytes (96\,KiB), which
matches the per-token size vLLM reports. A bank slot is therefore
$9.6\times$ smaller than a prompt token. Table~\ref{tab:measured_bank_sizes}
lists the measured bank tensors; each equals the slot count times 10\,KiB.
The serialized files are slightly larger: 2.085\,MiB for PHYSICS and
8.930\,MiB for the combined dialogue banks.

\begin{table}[!htbp]
\centering
\small
\setlength{\tabcolsep}{5pt}
\caption{Measured Qwen3 bank sizes (K and V tensors only, bf16).}
\label{tab:measured_bank_sizes}
\begin{tabular}{lrr}
\toprule
Bank & Slots & Size (MiB) \\
\midrule
PHYSICS subject & 195 & 1.904 \\
Dialogue target & 480 & 4.688 \\
Dialogue reference & 432 & 4.219 \\
Dialogue target + reference & 912 & 8.906 \\
\bottomrule
\end{tabular}
\end{table}

\paragraph{gpt-oss-120b estimates.}
Table~\ref{tab:gptoss_kv_supplement}
applies the same formulas: a prompt token occupies
$36\times8\times64\times2\times2$ bytes (72\,KiB), and a bank slot at
5 layers and 4 KV groups occupies $5\times4\times64\times2\times2$ bytes
(5\,KiB). Banks are content-matched: 
The prompt estimate assumes every layer stores every prompt token;
gpt-oss's windowed layers store fewer, so it is an upper bound.

\begin{table}[!htbp]
\centering
\small
\setlength{\tabcolsep}{6pt}
\caption{Estimated gpt-oss-120b storage for the reminder text.}
\label{tab:gptoss_kv_supplement}
\begin{tabular}{lrrr}
\toprule
Task & Prompt KV (MiB) & MI bank (MiB) & Ratio \\
\midrule
Self-evaluation & 53.0--59.3 & 3.1--3.5 & $\approx17\times$ \\
Generation & 27.8--29.5 & 1.6--1.8 & $\approx17\times$ \\
\bottomrule
\end{tabular}
\end{table}

\paragraph{Throughput.}
We ran 234 HARDMath examples per method (eight category/part
combinations) on Qwen3-30B-A3B with vLLM 0.11.2 on one H200, with
\texttt{max\_new\_tokens=16384} and identical launcher and batch settings.
Table~\ref{tab:hardmath_measured_throughput} reports total output tokens
divided by wall time.

\begin{table}[!htbp]
\centering
\small
\setlength{\tabcolsep}{6pt}
\caption{HARDMath generation time and throughput.}
\label{tab:hardmath_measured_throughput}
\begin{tabular}{lrrr}
\toprule
Method & Output tokens & Wall time (h) & Tokens/s \\
\midrule
Plain & 3,057,697 & 16.51 & 51.4 \\
Prompt & 2,217,066 & 13.27 & 46.4 \\
MI & 3,042,738 & 16.43 & 51.4 \\  
\bottomrule
\end{tabular}
\end{table}

MI and Plain have the same throughput and produce about the same amount of
text, so they take about the same time. Prompt produces about 27\% fewer
tokens and finishes sooner, despite lower throughput. Because output
lengths differ, this compares end-to-end cost, not the per-token overhead
of steering.

\section{Automated attention head/layer selection}
\label{app:selector}

The selector ranks attention units using calibration
diagnostics without task scores. Units are query heads
on Llama and KV groups on Qwen3 and gpt-oss; selected
groups expose all associated query heads to the bank.
Candidate layers are sampled every two or four layers.

\begin{algorithm}[H]
\caption{Layer and attention-unit selection}
\label{alg:selector}
\small
\textbf{Input:} candidate layers $\mathcal{L}_{\mathrm{cand}}$
and units $\mathcal{U}_\ell$, calibration prompts $\mathcal{C}$,
target and reference banks, budgets $m$ layers and $k$ units
per layer, and steering gains.\\
\textbf{Output:} frozen selector artifact $\mathcal{A}$.

\begin{enumerate}[leftmargin=1.7em,itemsep=0.25em,topsep=0.35em]
    \item Run calibration prompts and record alignment
    margins and attention assigned to each bank.

    \item Rank units within each layer by
    \begin{equation}
    U_{\ell,u}=\overline{a}_{\ell,u},
    \label{eq:head-selector}
    \end{equation}
    where $\overline{a}_{\ell,u}$ is the mean alignment
    margin from Eq.~\eqref{eq:alignment-margin}.
    Retain the top $k$ units, $\widehat{\mathcal{U}}_\ell$,
    breaking ties by mean total attention assigned to
    the target bank.

    \item Retain the top $m$ layers by
    \begin{equation}
    U_\ell=\sum_{u\in\widehat{\mathcal{U}}_\ell}U_{\ell,u}.
    \label{eq:layer-selector}
    \end{equation}

    \item Save selected sites, layer gains $\rho_\ell$,
    calibration settings, and diagnostics as $\mathcal{A}$.
    Freeze it before evaluation.
\end{enumerate}
\end{algorithm}

Calibration uses the evaluation benchmark pool and formatting; overlap with scored inputs is unverified. The calibration configuration for each run, the data used to tune gains, and the ranking rule without a reference bank remain unspecified.

\section{Datasets and evaluation}
\label{app:benchmarks}

Banks are frozen before evaluation. Construction examples used to distill heuristic cards are excluded from the scored pool. Appendix~\ref{app:memory_bank_examples} gives the bank roles, wrappers, and descriptor examples. Table~\ref{tab:evaluation_protocol} lists the evaluation sizes and settings.

\paragraph{Assets and licenses.}
Meta-Llama-3.1-8B-Instruct is used under the Llama 3.1 Community License, and Qwen3-30B-A3B under Apache 2.0, following their official model cards. The IPIP-50 questionnaire items are in the public domain. HARDMath, PHYSICS, GSM8K, and MMLU are used under the terms of their original public releases. The anonymous repository linked in the abstract points to the exact evaluation assets. GPT-4o-mini and GPT-5.5-thinking are accessed through the OpenAI API under the provider's service terms. GPT-4o-mini judges personality responses. GPT-5.5-thinking grades reasoning responses and generates heuristic text for the reasoning banks. We do not redistribute the API models.

\begin{table*}[!htbp]
\centering
\small
\setlength{\tabcolsep}{3pt}
\renewcommand{\arraystretch}{1.08}
\caption{\textbf{Reported evaluation sizes and settings.} Counts are corpus
sizes or table denominators, not counts after judge filtering.}
\label{tab:evaluation_protocol}
\begin{tabular}{P{0.23\textwidth}P{0.30\textwidth}P{0.40\textwidth}}
\toprule
Evaluation & Reported sample basis & Scoring and settings \\
\midrule
Big Five self-evaluation & IPIP-50 items, presented one at a time & 1--5 Likert responses, reverse-keyed items rescored before aggregation \\
Big Five generation & Prompt and scored-response counts not reported & GPT-4o-mini ratings on 1--5 scales \\
Dialogue shift & 68 primary cases and 204 stress cases, from 34 roots & GPT-4o-mini scores post-shift turns \\
Long-context persistence & 204 cases before judge filtering & 8k/16k/24k opposite-style prefills, GPT-4o-mini $P/T/Q$ scores on 1--10 scales \\
PHYSICS & Up to 100 questions per subject, actual subject counts not reported & GPT-5.5-thinking, three 0--5 grading passes per answer, thinking mode on/off \\
HARDMath accuracy & Category denominators total 362 & GPT-5.5-thinking plus human review, 0/0.5/1 credit \\
GSM8K / MMLU & 200 / 500 questions & Accuracy after persona steering \\
HARDMath timing & 234 examples per method & One H200, vLLM 0.11.2, \texttt{max\_new\_tokens=16384} \\
Wrapper ablation & IPIP-50, all five traits on Meta-Llama-3.1-8B-Instruct & Greedy decoding, fixed selector and steering gains \\
gpt-oss-120b personality & Sample counts not reported & DAS drift coefficient 1.5 (1 in the main text) \\
\bottomrule
\end{tabular}
\end{table*}

\subsection{Big Five self-evaluation}

This benchmark uses IPIP-50 Big Five items~\citep{goldberg2006ipip}. We steer the model toward a target trait, then present the questionnaire one item at a time and require a single 1--5 Likert response for each item. Reverse-keyed items are rescored before trait aggregation. A representative item from the evaluation pool is ``Am the life of the party.'' The target shift $\Delta_{\mathrm{target}}$ is the change in the steered trait relative to the plain model, and the non-target drift $D_{\mathrm{non\text{-}target}}$ is the mean absolute change over the other four traits.

\subsection{Big Five generation}

The generation benchmark uses the same target traits but measures them on free-form responses rather than questionnaire digits. A representative prompt is ``Write a brief reply to a person feeling uncertain before an interview.'' GPT-4o-mini judges each response. It scores target-trait presence, opposite-trait pressure, the five Big Five traits, and coherence on a 1--5 scale. The tables report the target-score shift, an off-target statistic $D$, and the contrastive margin. We could not recover the exact definition of $D$ from our records. It is nonzero for Plain, so it should not be read as change from Plain.

\subsection{Dialogue shift}

We construct the dialogue corpus as a set of multi-turn user trajectories rather than independent single-turn prompts. The corpus contains 34 scenario roots, each instantiated at 8, 16, and 24 turns with two branches per length, for $34 \times 3 \times 2 = 204$ cases. The roots cover emotionally loaded support cases, operational or crisis cases, skill-building and habit cases, and lower-stakes planning or creative cases. The 8-turn version establishes the premise, the 16-turn version adds a complication, and the 24-turn version adds a later-stage resolution or changed priority. The A and B branches share a root but vary concrete details so that success requires tracking the evolving user state.

The dialogue-shift benchmark uses the same cases to test whether steering can change behavior midway through an ongoing conversation. The primary tier uses the 68 cases with 8 turns. The stress tier uses all 204 cases. Each scenario root is assigned one of six trait transitions: \textbf{warm}$\rightarrow$\textbf{assertive}, \textbf{assertive}$\rightarrow$\textbf{warm}, \textbf{formal}$\rightarrow$\textbf{cautious}, \textbf{cautious}$\rightarrow$\textbf{formal}, \textbf{dismissive}$\rightarrow$\textbf{anxious}, or \textbf{anxious}$\rightarrow$\textbf{dismissive}. We inject a natural shift request at turn 5, 9, or 13 for the 8-, 16-, and 24-turn cases, respectively. One representative shift turn is ``Please switch gears. Be more anxious, worried, and uncertain from here on. I got one lukewarm reply and now I feel silly again. How should I recover?''

GPT-4o-mini scores only the post-shift assistant turns. Let $s_{\mathrm{target}}$ be the score of the requested new trait and $s_{\mathrm{old}}$ the score of the old trait. The primary adaptation metric is $\mathrm{Align}=s_{\mathrm{target}}-s_{\mathrm{old}}$. We also report the mean post-shift turn-quality score $Q_{\mathrm{turn}}$, defined as the average of usefulness, specificity, current-turn relevance, and non-genericness. The main comparison keeps the visible prompt and KV budget fixed after initialization. Prompt replacement uses more cache and is treated only as an audit condition.

\subsection{HARDMath}
\label{app:hardmath_benchmark}

HARDMath~\citep{fan2024hardmath} covers integral, Laplace-integral, ODE, polynomial-roots, and nondimensionalization problems. Table~\ref{tab:hardmath_memory_full} gives the category denominators, which total 362. We did not record the example IDs or the sampling rule that produced these denominators.

The original automatic scorer was too brittle for mathematically equivalent expressions. For example, it treated $\frac{1}{20}$ and $0.05$ as different answers. We therefore regrade with GPT-5.5-thinking plus human review. Each answer receives 0, 0.5, or 1 point. Full credit requires a correct boxed answer up to mathematical equivalence. Partial credit is assigned when the boxed answer is only partly complete but still captures a requested component of the official answer. For example, a problem may ask for both lower and upper asymptotic behavior, and the model gives only one asymptote.

The HARDMath banks contain process heuristics distilled from construction examples. Plain receives only the benchmark problem. Prompt places heuristic text in the visible context, and MI encodes heuristic text using the templates in Appendix~\ref{app:bank_construction_details}. We did not record whether Prompt and MI used exactly the same descriptor and variant in each run.

\subsection{PHYSICS}
\label{app:physics_benchmark}

The PHYSICS benchmark~\citep{feng2025physics} tests whether memory banks can carry reusable problem-solving heuristics rather than persona reminders. We use the text-only PHYSICS split. We evaluate up to 100 questions in each of six subjects: classical mechanics, quantum mechanics, thermodynamics and statistical mechanics, electromagnetism, atomic physics, and optics. Subject-specific heuristic cards are distilled from construction examples and then stripped of source question text, answer fragments, numerical constants, and solution-order traces before evaluation.

GPT-5.5-thinking grades each answer three times against a question-specific rubric, and we report the average normalized score. The rubric gives partial credit: a score of 5 means correct, 4 means mostly correct with a minor issue, 3 means substantial partial credit, 2 means some correct setup but major errors or incompleteness, 1 means very weak progress, and 0 means wrong or unusable. We evaluate with Qwen3 thinking mode off (no-CoT) and on (CoT).

\subsection{Long-context persistence}

Long-context persistence uses the same 204 dialogue cases as the shift benchmark. Here the stressor is a long opposite-style prefill instead of an explicit shift request. Before the task dialogue begins, we prefill the visible context with non-task assistant-history text written in the opposite style. The controller is then initialized once and must preserve the target style without repeatedly reinserting instructions into the visible transcript.

GPT-4o-mini scores every assistant turn on six style dimensions and four turn-quality dimensions. It also scores the full conversation on target-style persistence, coherence, user-state consistency, non-repetitiveness, and overall quality. Tables~\ref{tab:long_context_trait_budget_details} and~\ref{tab:llama_long_context_trait_budget_details} report $P$ (conversation-level target persistence), $T$ (mean turn-level target-style score), and $Q$ (conversation-level overall quality), all on 1--10 scales.

\subsection{GSM8K and MMLU}

We use GSM8K~\citep{cobbe2021gsm8k} and MMLU~\citep{hendrycks2021mmlu} only as coarse capability checks after steering. The Qwen3 runs use 200 GSM8K questions and 500 MMLU questions. We do not use these benchmarks to tune steering. They check whether persona steering causes an obvious loss of capability on standard reasoning and knowledge tasks.

\section{Memory-bank construction and examples}
\label{app:memory_bank_examples}
\label{app:bank_construction_details}
\label{app:rebuttal_additions}

Table~\ref{tab:memory_bank_construction_examples} summarizes the bank roles
and template variants. Equation~\eqref{eq:text-bank} specifies the residual
stream hook, the layer's input RMSNorm and K/V projections, and Qwen3 key
normalization. Only descriptor-token positions become slots. Keys are stored
before \RoPE{}.

\begin{table*}[!htbp]
\centering
\small
\setlength{\tabcolsep}{3pt}
\caption{\textbf{Memory-bank construction by task.} The template strings below
are reported separately from the descriptor examples. Complete per-run
descriptor versions and bank slot inventories remain unavailable.}
\label{tab:memory_bank_construction_examples}
\begin{tabular}{P{0.25\textwidth}P{0.39\textwidth}P{0.29\textwidth}}
\toprule
Task & Bank roles & Template variants \\
\midrule
Big Five questionnaire & Target trait, optional opposite-trait reference & Two wrappers listed, combined recipe unresolved \\
Big Five generation & Target trait, optional opposite-trait reference & Three wrapped variants \\
Persistence and dialogue shift & Target style, optional old/opposite-style reference & Three wrapped variants \\
PHYSICS & Subject target and auxiliary planning/completion banks, no reference & Two wrappers and identity \\
HARDMath & Category target and auxiliary planning/completion banks, no reference & Two wrappers and identity \\
\bottomrule
\end{tabular}
\end{table*}

The task-specific wrapper strings are listed below. The placeholder
\texttt{\{descriptor\}} is replaced by the bank text, and
\verb|\n| denotes an actual line break. Each variant is encoded
separately. Only descriptor-token positions are retained as slots.
The wrapper text affects their contextualized states but does not
itself become bank content.

\begin{lstlisting}[language={},frame=single,basicstyle=\ttfamily\footnotesize]
Big Five questionnaire:
Stable self-profile for questionnaire answers:\n{descriptor}
When deciding whether a statement describes you:\n{descriptor}

Big Five generation, long-context persistence, and dialogue shift:
Interaction style note:\n{descriptor}
Internal response style:\n{descriptor}
Behavioral steering note for replies:\n{descriptor}

PHYSICS:
Internal physics reasoning prior:\n{descriptor}
Hidden physics skill bank:\n{descriptor}
{descriptor}

HARDMath:
Internal applied mathematics reasoning prior:\n{descriptor}
Hidden HARDMath skill bank:\n{descriptor}
{descriptor}
\end{lstlisting}

A teacher model first distills the reasoning heuristics from construction
examples. Wrapping and encoding that text require no gradient updates to
the steered model.

The questionnaire wrapper list contains two strings, while the combined
679-slot ablation setting was described as using three templates. The third
template and the composition of that setting remain unresolved.
Appendix~\ref{app:wrapper_ablation} retains the reported ablation values.
The measured 195-slot PHYSICS bank and 480/432-slot dialogue target/reference
banks are documented in Appendix~\ref{app:measured_efficiency}. These counts
do not provide the full auxiliary-bank inventory for every run.

\paragraph{Exact implementation examples.}
The following descriptor and card spans are quoted from the implementation
and included with source identifiers and hashes in
\path{appendix/bank_examples.json}. They have not been matched to the
artifacts used for the reported benchmark rows. The HARDMath entry is one
card from the ODE bank. Exact PHYSICS descriptor spans remain unavailable.

\begin{lstlisting}[language={},frame=single,basicstyle=\ttfamily\footnotesize]
Base openness descriptor:
You are an open, imaginative assistant. Explore unusual but relevant angles, welcome new ideas, and sound intellectually curious without becoming random or impractical.
\end{lstlisting}

\begin{lstlisting}[language={},frame=single,basicstyle=\ttfamily\footnotesize]
Warm dialogue descriptor:
You are a warm, caring assistant. Sound friendly, supportive, emotionally attentive, and gently encouraging without becoming overly sentimental.
\end{lstlisting}

\begin{lstlisting}[language={},frame=single,basicstyle=\ttfamily\footnotesize]
HARDMath ODE card (ode_integrate):
Trigger: an ordinary differential equation gives a derivative of y
Core move: integrate with constants determined by all boundary or initial conditions
Anchor: integrate in the correct variable and apply each condition after the general antiderivative is available
Check: the final y should satisfy the differential equation and every condition
\end{lstlisting}

\section{Results breakdown by task}
\label{app:results_breakdown}

\subsection{Big Five self-evaluation}
\label{app:bigfive_self_eval_details}

\begin{table}[H]
\centering
\small
\setlength{\tabcolsep}{8pt}
\caption{\textbf{Big Five questionnaire self-evaluation on Meta-Llama-3.1-8B-Instruct.} Columns report target-trait score, target shift from the plain model, non-target drift, and drift-adjusted score, where $\mathrm{DAS}=\Delta_{\mathrm{target}}-D_{\mathrm{non\text{-}target}}$. Bold indicates the highest DAS among steering methods within each trait.}
\label{tab:bigfive_self_eval_details}
\begin{tabular}{llrrrr}
\toprule
Trait & Method & Score & Shift & Drift & DAS \\
\midrule
Openness & Plain & 3.5 & 0.0 & 0.000 & 0.000 \\
 & Prompt & 4.1 & +0.6 & 0.350 & 0.250 \\
 & CAA & 3.8 & +0.3 & 0.050 & 0.250 \\
 & Ours & 4.6 & +1.1 & 0.175 & \textbf{0.925} \\
\addlinespace[2pt]
Conscientiousness & Plain & 3.9 & 0.0 & 0.000 & 0.000 \\
 & Prompt & 4.2 & +0.3 & 0.625 & -0.325 \\
 & CAA & 4.3 & +0.4 & 0.025 & 0.375 \\
 & Ours & 4.9 & +1.1 & 0.100 & \textbf{1.000} \\
\addlinespace[2pt]
Extraversion & Plain & 2.8 & 0.0 & 0.000 & 0.000 \\
 & Prompt & 4.7 & +1.9 & 0.350 & \textbf{1.550} \\
 & CAA & 2.9 & +0.1 & 0.225 & -0.125 \\
 & Ours & 3.0 & +0.2 & 0.050 & 0.150 \\
\addlinespace[2pt]
Agreeableness & Plain & 4.7 & 0.0 & 0.000 & 0.000 \\
 & Prompt & 4.9 & +0.2 & 0.250 & -0.050 \\
 & CAA & 4.7 & +0.0 & 0.075 & -0.075 \\
 & Ours & 4.9 & +0.2 & 0.125 & \textbf{0.075} \\
\addlinespace[2pt]
Neuroticism & Plain & 2.0 & 0.0 & 0.000 & 0.000 \\
 & Prompt & 4.5 & +2.5 & 0.700 & \textbf{1.800} \\
 & CAA & 2.7 & +0.7 & 0.200 & 0.500 \\
 & Ours & 4.3 & +2.3 & 0.575 & 1.725 \\
\addlinespace[2pt]
Average & Plain & -- & 0.0 & 0.000 & 0.000 \\
 & Prompt & -- & +1.10 & 0.455 & 0.645 \\
 & CAA & -- & +0.30 & 0.115 & 0.185 \\
 & Ours & -- & +0.98 & 0.205 & \textbf{0.775} \\
\bottomrule
\end{tabular}
\end{table}

\begin{table}[H]
\centering
\small
\setlength{\tabcolsep}{8pt}
\caption{\textbf{Big Five questionnaire self-evaluation on Qwen3-30B-A3B.} Columns report target-trait score, target shift from the plain model, non-target drift, and drift-adjusted score, where $\mathrm{DAS}=\Delta_{\mathrm{target}}-D_{\mathrm{non\text{-}target}}$. Bold indicates the highest DAS among steering methods within each trait.}
\label{tab:bigfive_self_eval_details_qwen}
\begin{tabular}{llrrrr}
\toprule
Trait & Method & Score & Shift & Drift & DAS \\
\midrule
Openness & Plain & 4.1 & 0.0 & 0.000 & 0.000 \\
 & Prompt & 4.8 & +0.7 & 0.475 & 0.225 \\
 & CAA & 4.4 & +0.3 & 0.075 & 0.225 \\
 & Ours & 4.6 & +0.5 & 0.100 & \textbf{0.400} \\
\addlinespace[2pt]
Conscientiousness & Plain & 4.5 & 0.0 & 0.000 & 0.000 \\
 & Prompt & 4.9 & +0.4 & 0.525 & -0.125 \\
 & CAA & 4.5 & +0.1 & 0.075 & \textbf{0.025} \\
 & Ours & 4.6 & +0.1 & 0.150 & -0.050 \\
\addlinespace[2pt]
Extraversion & Plain & 3.6 & 0.0 & 0.000 & 0.000 \\
 & Prompt & 5.0 & +1.4 & 0.500 & \textbf{0.900} \\
 & CAA & 3.7 & +0.1 & 0.100 & 0.000 \\
 & Ours & 3.8 & +0.2 & 0.150 & 0.050 \\
\addlinespace[2pt]
Agreeableness & Plain & 4.2 & 0.0 & 0.000 & 0.000 \\
 & Prompt & 5.0 & +0.8 & 0.500 & 0.300 \\
 & CAA & 4.3 & +0.1 & 0.075 & 0.025 \\
 & Ours & 4.9 & +0.7 & 0.150 & \textbf{0.550} \\
\addlinespace[2pt]
Neuroticism & Plain & 2.5 & 0.0 & 0.000 & 0.000 \\
 & Prompt & 3.6 & +1.1 & 0.625 & \textbf{0.475} \\
 & CAA & 2.8 & +0.4 & 0.100 & 0.300 \\
 & Ours & 3.0 & +0.5 & 0.100 & 0.400 \\
\addlinespace[2pt]
Average & Plain & -- & 0.0 & 0.000 & 0.000 \\
 & Prompt & -- & +0.88 & 0.525 & \textbf{0.355} \\
 & CAA & -- & +0.20 & 0.085 & 0.115 \\
 & Ours & -- & +0.40 & 0.130 & 0.270 \\
\bottomrule
\end{tabular}
\end{table}

The per-trait tables clarify that the aggregate Pareto picture in Table~\ref{tab:core_tradeoff} is not uniform across all traits. Prompting remains strongest for the largest raw movement on extraversion and neuroticism, while memory-bank steering is strongest on several drift-adjusted rows, especially openness and agreeableness on Qwen3 and openness, conscientiousness, and agreeableness on Llama.

\subsection{Wrapper ablation}
\label{app:wrapper_ablation}

Table~\ref{tab:wrapper_ablation_supplement} reports an additional
Llama-3.1-8B IPIP-50 questionnaire evaluation over all five traits
with greedy decoding. Every row uses the same frozen per-trait
selector and steering hyperparameters. The slot-matched raw
condition repeats the descriptor without wrapper diversity.
The raw and combined conditions have the same mean target shift,
0.98, with mean drift of 0.217 and 0.205, respectively.
This comparison covers one backbone and one task.

\begin{table}[H]
\centering
\small
\setlength{\tabcolsep}{5pt}
\caption{\textbf{Wrapper ablation on Llama-3.1-8B.}
All conditions use the same frozen selector and steering
hyperparameters. The final column gives the score reported for
this experiment.}
\label{tab:wrapper_ablation_supplement}
\begin{tabular}{lrrrr}
\toprule
Condition & Slots & Mean shift & Mean drift & Reported score \\
\midrule
Raw descriptor & 226 & 0.98 & 0.217 & 0.655 \\
Raw descriptor repeated & 679 & 0.98 & 0.233 & 0.631 \\
Wrapper 1 only & 226 & 0.96 & 0.205 & 0.653 \\
Wrapper 2 only & 226 & 0.96 & 0.195 & 0.668 \\
Combined setting & 679 & 0.98 & 0.205 & 0.673 \\
\bottomrule
\end{tabular}
\end{table}

\subsection{Big Five generation}
\label{app:bigfive_generation_details}

\begin{table}[H]
\centering
\small
\setlength{\tabcolsep}{7pt}
\caption{\textbf{Big Five contrastive generation judged by an LLM judge on Meta-Llama-3.1-8B-Instruct.} Columns report target-trait score, target-score shift from the plain model, the reported off-target statistic $D$, contrastive margin, and generation-adjusted score, where $\mathrm{GAS}=\Delta_{\mathrm{target}}-D_{\mathrm{non\text{-}target}}$. The definition of $D$ remains unresolved, as discussed in Appendix~\ref{app:benchmarks}. Contrastive margin is a separate diagnostic.}
\label{tab:bigfive_generation_details}
\begin{tabular}{llrrrrr}
\toprule
Trait & Method & Score & Shift & $D$ & Margin & GAS \\
\midrule
Openness & Plain & 4.54 & +0.00 & 1.11 & 3.33 & -1.110 \\
 & Prompt & 4.79 & +0.25 & 1.04 & 3.62 & \textbf{-0.790} \\
 & CAA & 4.62 & +0.08 & 1.10 & 3.29 & -1.020 \\
 & Ours & 4.50 & -0.04 & 1.07 & 3.12 & -1.110 \\
\addlinespace[2pt]
Conscientiousness & Plain & 4.83 & +0.00 & 0.68 & 3.71 & -0.680 \\
 & Prompt & 4.79 & -0.04 & 0.64 & 3.62 & -0.680 \\
 & CAA & 4.92 & +0.09 & 0.66 & 3.88 & -0.570 \\
 & Ours & 4.96 & +0.13 & 0.69 & 3.92 & \textbf{-0.560} \\
\addlinespace[2pt]
Extraversion & Plain & 3.50 & +0.00 & 1.18 & 1.38 & -1.180 \\
 & Prompt & 4.33 & +0.83 & 1.25 & 2.71 & \textbf{-0.420} \\
 & CAA & 3.92 & +0.42 & 1.18 & 1.83 & -0.760 \\
 & Ours & 4.12 & +0.62 & 1.19 & 2.29 & -0.570 \\
\addlinespace[2pt]
Agreeableness & Plain & 4.83 & +0.00 & 0.96 & 3.79 & -0.960 \\
 & Prompt & 4.92 & +0.09 & 0.55 & 3.83 & \textbf{-0.460} \\
 & CAA & 4.88 & +0.05 & 1.02 & 3.83 & -0.970 \\
 & Ours & 4.92 & +0.09 & 0.95 & 3.88 & -0.860 \\
\addlinespace[2pt]
Neuroticism & Plain & 3.62 & +0.00 & 0.99 & 1.12 & -0.990 \\
 & Prompt & 4.29 & +0.67 & 0.94 & 2.42 & -0.270 \\
 & CAA & 3.50 & -0.12 & 0.93 & 1.00 & -1.050 \\
 & Ours & 4.42 & +0.80 & 0.71 & 2.79 & \textbf{0.090} \\
\addlinespace[2pt]
Average & Plain & 4.26 & +0.00 & 0.98 & 2.67 & -0.984 \\
 & Prompt & 4.62 & +0.36 & 0.88 & 3.24 & \textbf{-0.524} \\
 & CAA & 4.37 & +0.11 & 0.98 & 2.77 & -0.874 \\
 & Ours & 4.58 & +0.32 & 0.92 & 3.20 & -0.602 \\
\bottomrule
\end{tabular}
\end{table}

\begin{table}[H]
\centering
\small
\setlength{\tabcolsep}{7pt}
\caption{\textbf{Big Five contrastive generation judged by an LLM judge on Qwen3-30B-A3B.} Columns report target-trait score, target-score shift from the plain model, the reported off-target statistic $D$, contrastive margin, and generation-adjusted score, where $\mathrm{GAS}=\Delta_{\mathrm{target}}-D_{\mathrm{non\text{-}target}}$. The definition of $D$ remains unresolved, as discussed in Appendix~\ref{app:benchmarks}.}
\label{tab:bigfive_generation_details_qwen}
\begin{tabular}{llrrrrr}
\toprule
Trait & Method & Score & Shift & $D$ & Margin & GAS \\
\midrule
Openness & Plain & 2.17 & +0.00 & 0.83 & 0.67 & -0.830 \\
 & Prompt & 2.83 & +0.66 & 0.83 & 1.67 & -0.170 \\
 & CAA & 1.83 & -0.34 & 1.00 & -0.33 & -1.340 \\
 & Ours & 3.83 & +1.66 & 1.12 & 2.00 & \textbf{0.540} \\
\addlinespace[2pt]
Conscientiousness & Plain & 3.83 & +0.00 & 0.71 & 2.83 & -0.710 \\
 & Prompt & 4.17 & +0.34 & 0.58 & 3.17 & -0.240 \\
 & CAA & 3.83 & +0.00 & 0.62 & 2.83 & -0.620 \\
 & Ours & 5.00 & +1.17 & 0.92 & 4.00 & \textbf{0.250} \\
\addlinespace[2pt]
Extraversion & Plain & 2.00 & +0.00 & 0.92 & 0.83 & -0.920 \\
 & Prompt & 3.67 & +1.67 & 0.92 & 2.67 & \textbf{0.750} \\
 & CAA & 2.00 & +0.00 & 0.96 & 0.83 & -0.960 \\
 & Ours & 3.33 & +1.33 & 1.17 & 0.83 & 0.160 \\
\addlinespace[2pt]
Agreeableness & Plain & 3.50 & +0.00 & 0.62 & 2.50 & -0.620 \\
 & Prompt & 4.17 & +0.67 & 0.71 & 3.17 & -0.040 \\
 & CAA & 3.50 & +0.00 & 0.75 & 2.33 & -0.750 \\
 & Ours & 5.00 & +1.50 & 0.88 & 4.00 & \textbf{0.620} \\
\addlinespace[2pt]
Neuroticism & Plain & 1.00 & +0.00 & 0.96 & -4.00 & -0.960 \\
 & Prompt & 2.67 & +1.67 & 0.67 & -0.33 & \textbf{1.000} \\
 & CAA & 1.00 & +0.00 & 0.96 & -4.00 & -0.960 \\
 & Ours & 2.67 & +1.67 & 1.04 & -0.67 & 0.630 \\
\addlinespace[2pt]
Average & Plain & 2.50 & +0.00 & 0.81 & 0.57 & -0.808 \\
 & Prompt & 3.50 & +1.00 & 0.74 & 2.07 & 0.260 \\
 & CAA & 2.43 & -0.07 & 0.86 & 0.33 & -0.926 \\
 & Ours & 3.97 & +1.47 & 1.02 & 2.03 & \textbf{0.440} \\
\bottomrule
\end{tabular}
\end{table}

The generation tables show why the main-text claim is improvement over CAA rather than universal dominance over prompting. Memory-bank steering improves over CAA on the average Llama and Qwen3 generation metrics, but prompt remains the strongest visible-context baseline on several drift-adjusted rows, especially on Qwen3.

\subsection{Long-context persistence}

\begin{table}[H]
\centering
\small
\setlength{\tabcolsep}{2pt}
\caption{\textbf{Qwen3 long-context dialogue persistence by trait and prefill length.} Each method cell reports $P/T/Q$: conversation-level target persistence, mean turn-level target style score, and conversation-level overall quality. Prefill is non-task assistant-history context inserted before the dialogue. Prompt is the once-initialized visible-instruction baseline; Ours is memory-bank steering. Missing cells indicate no valid judged run after judge-completion filtering.}
\label{tab:long_context_trait_budget_details}
\begin{tabular}{llcccc}
\toprule
Trait & Prefill & Plain & Prompt & CAA & Ours \\
\midrule
Warm & 8k & 9.04\,/\,8.68\,/\,9.19 & 9.19\,/\,8.95\,/\,9.34 & 8.94\,/\,8.41\,/\,9.02 & 8.51\,/\,8.29\,/\,8.39 \\
 & 16k & 9.01\,/\,8.62\,/\,9.14 & 9.12\,/\,8.79\,/\,9.21 & 8.91\,/\,8.43\,/\,9.00 & 8.90\,/\,8.43\,/\,8.92 \\
 & 24k & 8.96\,/\,8.64\,/\,9.10 & 9.06\,/\,8.73\,/\,9.13 & 8.94\,/\,8.48\,/\,8.99 & 8.91\,/\,8.47\,/\,9.00 \\
\cmidrule(lr){1-6}
Formal & 8k & 8.07\,/\,6.31\,/\,9.07 & 8.41\,/\,6.68\,/\,9.18 & 8.06\,/\,6.43\,/\,9.01 & 8.04\,/\,6.40\,/\,8.97 \\
 & 16k & 8.04\,/\,6.36\,/\,9.01 & 8.16\,/\,6.58\,/\,9.05 & 8.10\,/\,6.51\,/\,8.98 & 8.06\,/\,6.43\,/\,9.00 \\
 & 24k & 8.05\,/\,6.35\,/\,9.04 & 8.18\,/\,6.53\,/\,9.05 & 8.05\,/\,6.47\,/\,8.97 & 8.05\,/\,6.41\,/\,8.98 \\
\cmidrule(lr){1-6}
Assertive & 8k & 8.94\,/\,7.88\,/\,9.24 & 9.01\,/\,8.12\,/\,9.30 & 8.79\,/\,7.59\,/\,9.01 & 8.73\,/\,7.56\,/\,9.01 \\
 & 16k & 8.83\,/\,7.91\,/\,9.05 & 9.01\,/\,8.15\,/\,9.19 & 8.77\,/\,7.60\,/\,8.98 & 8.72\,/\,7.57\,/\,8.99 \\
& 24k & 8.85\,/\,7.91\,/\,9.05 & 8.98\,/\,8.06\,/\,9.11 & 8.73\,/\,7.62\,/\,8.97 & 8.71\,/\,7.57\,/\,8.96 \\

\cmidrule(lr){1-6}
Cautious & 8k & 8.65\,/\,6.18\,/\,9.12 & 8.86\,/\,6.34\,/\,9.20 & 8.65\,/\,6.23\,/\,9.02 & 8.71\,/\,6.29\,/\,8.97 \\
 & 16k & 8.60\,/\,6.13\,/\,9.10 & 8.72\,/\,6.25\,/\,9.12 & 8.66\,/\,6.12\,/\,8.99 & 8.74\,/\,6.19\,/\,8.99 \\
 & 24k & 8.56\,/\,6.10\,/\,9.06 & 8.65\,/\,6.19\,/\,9.06 & 8.67\,/\,6.10\,/\,8.99 & 8.70\,/\,6.22\,/\,8.99 \\
\cmidrule(lr){1-6}
Dismissive & 8k & 1.00\,/\,1.01\,/\,9.69 & 1.61\,/\,1.26\,/\,9.49 & 1.00\,/\,1.04\,/\,9.28 & 2.35\,/\,1.99\,/\,6.56 \\
 & 16k & 1.00\,/\,1.02\,/\,9.54 & 1.05\,/\,1.05\,/\,9.60 & 1.01\,/\,1.03\,/\,9.17 & 1.85\,/\,1.84\,/\,6.51 \\
 & 24k & 1.00\,/\,1.01\,/\,9.58 & 1.00\,/\,1.02\,/\,9.58 & 1.00\,/\,1.04\,/\,9.23 & 1.68\,/\,1.60\,/\,6.55 \\
\cmidrule(lr){1-6}
Anxious & 8k & 4.96\,/\,1.67\,/\,8.90 & 5.59\,/\,2.14\,/\,8.90 & 5.73\,/\,1.96\,/\,8.55 & 6.23\,/\,2.54\,/\,6.53 \\
& 16k & 4.99\,/\,1.74\,/\,8.73 & 4.94\,/\,1.73\,/\,8.81 & 5.86\,/\,2.01\,/\,8.52 & 6.21\,/\,2.39\,/\,7.12 \\
& 24k & 4.69\,/\,1.75\,/\,8.75 & 4.88\,/\,1.74\,/\,8.77 & 6.05\,/\,2.03\,/\,8.51 & 6.27\,/\,2.27\,/\,7.35 \\
\bottomrule
\end{tabular}
\end{table}

\begin{table}[H]
\centering
\small
\setlength{\tabcolsep}{2pt}
\caption{\textbf{Meta-Llama-3.1-8B-Instruct long-context dialogue persistence by trait and prefill length.} Each cell reports $P/T/Q$: conversation-level target persistence, mean turn-level target style score, and conversation-level overall quality. The main text reports only the model-level averages; this table gives the full trait$\times$budget breakdown. Plain is the opposite-prefill baseline. Prompt adds one visible target-style system instruction before the same opposite-prefill context. CAA is the frozen post-attention residual baseline (layer 14, scale 2), and Ours reports the frozen memory-bank configuration used in the final judged audit.}
\label{tab:llama_long_context_trait_budget_details}
\begin{tabular}{llcccc}
\toprule
Trait & Prefill & Plain & Prompt & CAA & Ours \\
\midrule
Warm & 8k & 7.66\,/\,7.38\,/\,8.57 & 8.76\,/\,8.06\,/\,8.99 & 8.23\,/\,7.49\,/\,8.83 & 8.08\,/\,7.41\,/\,8.83 \\
  & 16k & 8.17\,/\,7.52\,/\,8.83 & 8.75\,/\,8.00\,/\,8.95 & 7.92\,/\,7.47\,/\,8.71 & 8.26\,/\,7.43\,/\,8.83 \\
  & 24k & 7.72\,/\,7.60\,/\,8.61 & 8.70\,/\,8.01\,/\,9.00 & 8.29\,/\,7.71\,/\,8.81 & 7.92\,/\,7.46\,/\,8.70 \\
\cmidrule(lr){1-6}
Formal & 8k & 7.66\,/\,6.09\,/\,8.79 & 8.01\,/\,6.20\,/\,8.98 & 8.04\,/\,6.56\,/\,9.00 & 7.69\,/\,6.01\,/\,8.76 \\
  & 16k & 7.51\,/\,6.05\,/\,8.75 & 7.96\,/\,6.27\,/\,8.93 & 8.11\,/\,6.58\,/\,9.00 & 7.51\,/\,6.01\,/\,8.63 \\
  & 24k & 7.48\,/\,6.04\,/\,8.67 & 7.93\,/\,6.26\,/\,8.93 & 8.06\,/\,6.62\,/\,8.99 & 7.52\,/\,5.99\,/\,8.64 \\
\cmidrule(lr){1-6}
Assertive & 8k & 7.80\,/\,6.79\,/\,8.78 & 7.91\,/\,7.10\,/\,8.82 & 8.32\,/\,7.25\,/\,8.98 & 7.54\,/\,6.37\,/\,8.51 \\
  & 16k & 6.55\,/\,6.26\,/\,7.87 & 7.68\,/\,6.95\,/\,8.58 & 8.12\,/\,7.22\,/\,8.89 & 6.95\,/\,6.25\,/\,8.00 \\
  & 24k & 7.10\,/\,6.50\,/\,8.19 & 7.67\,/\,6.89\,/\,8.61 & 7.99\,/\,7.21\,/\,8.87 & 7.04\,/\,6.21\,/\,7.97 \\
\cmidrule(lr){1-6}
Cautious & 8k & 7.36\,/\,5.73\,/\,8.46 & 8.39\,/\,6.39\,/\,9.00 & 7.84\,/\,6.17\,/\,8.77 & 7.46\,/\,5.71\,/\,8.48 \\
  & 16k & 7.47\,/\,5.70\,/\,8.62 & 8.32\,/\,6.37\,/\,8.97 & 7.92\,/\,6.16\,/\,8.86 & 7.34\,/\,5.70\,/\,8.34 \\
  & 24k & 7.47\,/\,5.85\,/\,8.57 & 8.22\,/\,6.30\,/\,8.97 & 7.98\,/\,6.28\,/\,8.80 & 7.87\,/\,5.86\,/\,8.78 \\
\cmidrule(lr){1-6}
Dismissive & 8k & 1.00\,/\,1.06\,/\,8.69 & 7.62\,/\,6.06\,/\,5.87 & 1.02\,/\,1.19\,/\,8.74 & 3.63\,/\,2.74\,/\,4.45 \\
  & 16k & 1.01\,/\,1.06\,/\,8.71 & 8.41\,/\,7.09\,/\,5.39 & 1.10\,/\,1.20\,/\,8.66 & 4.08\,/\,2.69\,/\,4.14 \\
  & 24k & 1.03\,/\,1.07\,/\,8.73 & 8.45\,/\,6.87\,/\,5.37 & 1.21\,/\,1.16\,/\,8.57 & 3.92\,/\,2.62\,/\,4.46 \\
\cmidrule(lr){1-6}
Anxious & 8k & 5.88\,/\,2.09\,/\,8.06 & 6.11\,/\,2.20\,/\,8.04 & 6.08\,/\,2.29\,/\,8.03 & 6.53\,/\,2.72\,/\,7.78 \\
  & 16k & 6.09\,/\,2.14\,/\,8.07 & 6.56\,/\,2.88\,/\,8.07 & 6.13\,/\,2.29\,/\,8.07 & 6.49\,/\,2.67\,/\,7.78 \\
  & 24k & 6.12\,/\,2.14\,/\,8.03 & 6.46\,/\,2.47\,/\,8.13 & 6.16\,/\,2.29\,/\,8.04 & 6.50\,/\,2.66\,/\,7.89 \\
\bottomrule
\end{tabular}
\end{table}

Qwen3 shows the clearest positive long-context cases under the budget-matched setup, especially on the more difficult low-baseline traits. Llama behaves differently: prompt remains the strongest raw persistence mechanism on average, while the latent methods trade off persistence against judged quality more sharply.

\subsection{Dialogue shift}

For readability, the appendix tables below foreground the two primary post-shift outcomes: adaptation alignment $\mathrm{Align}=s_{\mathrm{target}}-s_{\mathrm{old}}$ and judged turn quality $Q_{\mathrm{turn}}$. The raw target-style and old-style component scores are omitted from the paper-level tables and remain available in the released results.

\begin{table}[H]
\centering
\small
\setlength{\tabcolsep}{4pt}
\renewcommand{\arraystretch}{1.06}
\caption{\textbf{Qwen3 dialogue-shift post-shift alignment by transition and dialogue length.} Entries report $\mathrm{Align}$ on post-shift turns. Bold marks the best matched-budget method in each row among Plain, Prompt, CAA, and Ours. Prompt-repl.\ is shown only as an audit because it adds visible steering text at the shift point.}
\label{tab:dialogue_shift_transition_length_details}
\begin{tabular}{P{0.24\textwidth}c cccccc}
\toprule
Transition & Turns & Metric & Plain & Prompt & CAA & Ours & Prompt-repl. \\
\midrule
Warm$\rightarrow$Assertive & 8 & $\mathrm{Align}$ & -0.04 & -0.46 & \textbf{0.19} & -0.50 & -0.29 \\
Warm$\rightarrow$Assertive & 16 & $\mathrm{Align}$ & -0.23 & -0.25 & \textbf{0.00} & -0.34 & -0.29 \\
Warm$\rightarrow$Assertive & 24 & $\mathrm{Align}$ & -0.40 & -0.43 & \textbf{-0.30} & -0.80 & -0.35 \\
\midrule
Assertive$\rightarrow$Warm & 8 & $\mathrm{Align}$ & 1.33 & 1.17 & 1.56 & \textbf{1.88} & 1.12 \\
Assertive$\rightarrow$Warm & 16 & $\mathrm{Align}$ & 1.43 & 1.38 & 1.66 & \textbf{2.29} & 1.34 \\
Assertive$\rightarrow$Warm & 24 & $\mathrm{Align}$ & 1.25 & 1.24 & 1.38 & \textbf{1.90} & 1.17 \\
\midrule
Formal$\rightarrow$Cautious & 8 & $\mathrm{Align}$ & 0.22 & -0.06 & \textbf{0.46} & 0.06 & 0.11 \\
Formal$\rightarrow$Cautious & 16 & $\mathrm{Align}$ & -0.14 & -0.21 & \textbf{0.19} & -0.02 & -0.12 \\
Formal$\rightarrow$Cautious & 24 & $\mathrm{Align}$ & -0.29 & -0.44 & -0.17 & \textbf{-0.15} & -0.06 \\
\midrule
Cautious$\rightarrow$Formal & 8 & $\mathrm{Align}$ & 1.50 & 1.58 & \textbf{1.69} & 1.42 & 1.67 \\
Cautious$\rightarrow$Formal & 16 & $\mathrm{Align}$ & 0.79 & 0.89 & \textbf{1.16} & 0.71 & 1.21 \\
Cautious$\rightarrow$Formal & 24 & $\mathrm{Align}$ & 0.82 & 1.05 & \textbf{1.29} & 0.93 & 1.09 \\
\midrule
Dismissive$\rightarrow$Anxious & 8 & $\mathrm{Align}$ & 0.83 & 1.00 & 1.96 & \textbf{2.83} & 2.06 \\
Dismissive$\rightarrow$Anxious & 16 & $\mathrm{Align}$ & 1.02 & 1.55 & 1.65 & \textbf{2.74} & 4.17 \\
Dismissive$\rightarrow$Anxious & 24 & $\mathrm{Align}$ & 1.11 & 1.02 & \textbf{2.52} & 2.48 & 2.86 \\
\midrule
Anxious$\rightarrow$Dismissive & 8 & $\mathrm{Align}$ & -1.17 & -0.67 & \textbf{0.12} & -1.17 & 0.00 \\
Anxious$\rightarrow$Dismissive & 16 & $\mathrm{Align}$ & -0.93 & -0.79 & -0.44 & \textbf{-0.36} & 0.21 \\
Anxious$\rightarrow$Dismissive & 24 & $\mathrm{Align}$ & -0.77 & -0.55 & -0.42 & \textbf{-0.32} & -0.86 \\
\bottomrule
\end{tabular}
\end{table}

\begin{table}[H]
\centering
\small
\setlength{\tabcolsep}{4pt}
\renewcommand{\arraystretch}{1.06}
\caption{\textbf{Qwen3 dialogue-shift post-shift turn quality by transition and dialogue length.} Entries report $Q_{\mathrm{turn}}$ on post-shift turns. Bold marks the best matched-budget method in each row among Plain, Prompt, CAA, and Ours. Prompt-repl.\ is shown only as an audit because it adds visible steering text at the shift point.}
\label{tab:dialogue_shift_qturn_details}
\begin{tabular}{P{0.24\textwidth}c cccccc}
\toprule
Transition & Turns & Metric & Plain & Prompt & CAA & Ours & Prompt-repl. \\
\midrule
Warm$\rightarrow$Assertive & 8 & $Q_{\mathrm{turn}}$ & 8.67 & \textbf{8.72} & 8.64 & 7.86 & 8.65 \\
Warm$\rightarrow$Assertive & 16 & $Q_{\mathrm{turn}}$ & 8.73 & \textbf{8.77} & 8.71 & 7.64 & 8.78 \\
Warm$\rightarrow$Assertive & 24 & $Q_{\mathrm{turn}}$ & \textbf{8.70} & 8.66 & 8.59 & 7.25 & 8.63 \\
\midrule
Assertive$\rightarrow$Warm & 8 & $Q_{\mathrm{turn}}$ & 8.47 & \textbf{8.62} & 8.41 & 7.27 & 8.58 \\
Assertive$\rightarrow$Warm & 16 & $Q_{\mathrm{turn}}$ & \textbf{8.58} & 8.55 & 8.45 & 6.45 & 8.51 \\
Assertive$\rightarrow$Warm & 24 & $Q_{\mathrm{turn}}$ & \textbf{8.55} & 8.53 & 8.47 & 6.44 & 8.57 \\
\midrule
Formal$\rightarrow$Cautious & 8 & $Q_{\mathrm{turn}}$ & \textbf{8.60} & 8.58 & 8.51 & 8.44 & 8.49 \\
Formal$\rightarrow$Cautious & 16 & $Q_{\mathrm{turn}}$ & \textbf{8.55} & 8.50 & 8.54 & 8.27 & 8.49 \\
Formal$\rightarrow$Cautious & 24 & $Q_{\mathrm{turn}}$ & 8.39 & \textbf{8.43} & 8.35 & 8.19 & 8.40 \\
\midrule
Cautious$\rightarrow$Formal & 8 & $Q_{\mathrm{turn}}$ & 8.46 & 8.46 & \textbf{8.50} & 7.67 & 8.42 \\
Cautious$\rightarrow$Formal & 16 & $Q_{\mathrm{turn}}$ & 8.58 & \textbf{8.63} & 8.59 & 7.47 & 8.59 \\
Cautious$\rightarrow$Formal & 24 & $Q_{\mathrm{turn}}$ & 8.66 & \textbf{8.70} & 8.68 & 7.89 & 8.61 \\
\midrule
Dismissive$\rightarrow$Anxious & 8 & $Q_{\mathrm{turn}}$ & \textbf{8.56} & 8.49 & 8.33 & 6.28 & 7.53 \\
Dismissive$\rightarrow$Anxious & 16 & $Q_{\mathrm{turn}}$ & \textbf{8.56} & 8.24 & 8.31 & 4.79 & 6.27 \\
Dismissive$\rightarrow$Anxious & 24 & $Q_{\mathrm{turn}}$ & 8.44 & \textbf{8.52} & 7.42 & 3.98 & 6.99 \\
\midrule
Anxious$\rightarrow$Dismissive & 8 & $Q_{\mathrm{turn}}$ & \textbf{8.62} & 8.29 & 8.03 & 7.33 & 7.50 \\
Anxious$\rightarrow$Dismissive & 16 & $Q_{\mathrm{turn}}$ & \textbf{8.18} & 7.95 & 7.72 & 5.27 & 6.64 \\
Anxious$\rightarrow$Dismissive & 24 & $Q_{\mathrm{turn}}$ & \textbf{8.15} & 8.12 & 8.05 & 4.16 & 7.70 \\
\bottomrule
\end{tabular}
\end{table}

\begin{table}[H]
\centering
\small
\setlength{\tabcolsep}{4pt}
\renewcommand{\arraystretch}{1.06}
\caption{\textbf{Meta-Llama-3.1-8B-Instruct halfway-shift dialogue summary.} Each breakdown is split into two metric rows: post-shift alignment $\mathrm{Align}$ and judged turn quality $Q_{\mathrm{turn}}$. Bold marks the best method in each metric row. CAA is the frozen layer-14, scale-2 post-attention residual baseline; Ours is memory-bank summary update.}
\label{tab:llama_dialogue_shift_details}
\begin{tabular}{P{0.30\textwidth}P{0.12\textwidth}cccc}
\toprule
Breakdown & Metric & Plain & Prompt & CAA & Ours \\
\midrule
Overall & $\mathrm{Align}$ & 1.112 & 1.178 & \textbf{1.554} & 1.516 \\
 & $Q_{\mathrm{turn}}$ & \textbf{7.72} & 7.58 & 7.34 & 7.00 \\
\midrule
Anxious$\rightarrow$Dismissive & $\mathrm{Align}$ & 1.654 & 2.212 & \textbf{2.808} & 2.692 \\
 & $Q_{\mathrm{turn}}$ & \textbf{5.92} & 5.49 & 4.63 & 4.49 \\
Assertive$\rightarrow$Warm & $\mathrm{Align}$ & 1.026 & 1.087 & \textbf{1.209} & 1.148 \\
 & $Q_{\mathrm{turn}}$ & 8.30 & \textbf{8.35} & 8.23 & 8.03 \\
Cautious$\rightarrow$Formal & $\mathrm{Align}$ & 0.835 & 0.852 & \textbf{1.157} & 0.913 \\
 & $Q_{\mathrm{turn}}$ & 8.33 & \textbf{8.35} & 8.27 & 7.87 \\
Dismissive$\rightarrow$Anxious & $\mathrm{Align}$ & 3.615 & 3.240 & 4.298 & \textbf{4.577} \\
 & $Q_{\mathrm{turn}}$ & \textbf{7.06} & 6.44 & 5.98 & 5.85 \\
Formal$\rightarrow$Cautious & $\mathrm{Align}$ & -0.252 & \textbf{-0.113} & -0.191 & -0.209 \\
 & $Q_{\mathrm{turn}}$ & 8.11 & \textbf{8.14} & 8.05 & 7.77 \\
Warm$\rightarrow$Assertive & $\mathrm{Align}$ & 0.087 & 0.087 & \textbf{0.426} & 0.383 \\
 & $Q_{\mathrm{turn}}$ & 8.40 & 8.40 & \textbf{8.48} & 7.65 \\
\midrule
8 turns & $\mathrm{Align}$ & 1.185 & 1.204 & \textbf{1.398} & 1.278 \\
 & $Q_{\mathrm{turn}}$ & \textbf{7.84} & 7.57 & 7.52 & 7.22 \\
16 turns & $\mathrm{Align}$ & 1.373 & 1.429 & \textbf{1.833} & 1.829 \\
 & $Q_{\mathrm{turn}}$ & \textbf{7.56} & 7.36 & 7.15 & 6.88 \\
24 turns & $\mathrm{Align}$ & 0.873 & 0.964 & \textbf{1.380} & 1.344 \\
 & $Q_{\mathrm{turn}}$ & \textbf{7.82} & 7.76 & 7.43 & 7.03 \\
\bottomrule
\end{tabular}
\end{table}

The shift tables show the main matched-budget pattern from the paper: Qwen3 provides the clearest positive cases for latent behavior updates, especially on difficult transitions such as \texttt{dismissive}$\rightarrow$\texttt{anxious}, while Llama shows a sharper quality trade-off even when alignment improves.

\subsection{PHYSICS results}

\begin{table}[H]
\centering
\small
\setlength{\tabcolsep}{6pt}
\caption{\textbf{Full PHYSICS subject$\times$mode matrix on Qwen3-30B-A3B.} Values are mean 0--5 partial-credit judge scores, normalized to percent and averaged over three grading passes. Chain of thought (CoT) denotes Qwen3 thinking mode; no-CoT disables it.}
\label{tab:physics_memory_full}
\begin{tabular}{P{0.39\textwidth}lccc}
\toprule
Domain & Mode & Plain (\%) $\uparrow$ & Prompt (\%) $\uparrow$ & Ours (\%) $\uparrow$ \\
\midrule
Classical mechanics & no-CoT & 66.0 & \textbf{69.4} & 68.2 \\
Classical mechanics & CoT & 82.0 & 84.0 & \textbf{85.4} \\
Quantum mechanics & no-CoT & 65.4 & 63.5 & \textbf{68.1} \\
Quantum mechanics & CoT & 69.2 & 75.0 & \textbf{76.6} \\
Atomic physics & no-CoT & 71.4 & 71.8 & \textbf{73.2} \\
Atomic physics & CoT & 80.2 & 79.5 & \textbf{84.2} \\
Optics & no-CoT & 69.2 & 72.9 & \textbf{74.2} \\
Optics & CoT & \textbf{85.0} & 83.2 & 83.0 \\
Electromagnetism & no-CoT & 60.6 & 63.6 & \textbf{64.4} \\
Electromagnetism & CoT & 75.8 & 76.6 & \textbf{77.8} \\
Thermodynamics/statistical mechanics & no-CoT & 72.2 & 76.4 & \textbf{76.8} \\
Thermodynamics/statistical mechanics & CoT & 89.4 & 89.0 & \textbf{90.6} \\
\bottomrule
\end{tabular}
\end{table}

The PHYSICS results support the structured-heuristic use case: memory banks are most helpful when the benchmark rewards reusable subject-level process reminders rather than persona changes. Gains are especially visible in CoT mode, where the latent heuristic cards can guide longer reasoning traces without occupying visible prompt budget.

\subsection{HARDMath results}

\begin{table}[H]
\centering
\small
\setlength{\tabcolsep}{4pt}
\caption{\textbf{Full HARDMath category breakdown on Qwen3-30B-A3B with chain of thought.} Values are category partial-credit total / maximum partial-credit total, with the normalized percentage in parentheses.}
\label{tab:hardmath_memory_full}
\begin{tabular*}{\textwidth}{@{\extracolsep{\fill}}lccc@{}}
\toprule
Category & Plain & Prompt & Ours \\
\midrule
Integral 0 & 13.5 / 30 (45.0) & \textbf{14.5 / 30 (48.3)} & \textbf{14.5 / 30 (48.3)} \\
Integral 1 & \textbf{13.5 / 30 (45.0)} & 13.0 / 30 (43.3) & \textbf{13.5 / 30 (45.0)} \\
Laplace-integral 0 & 26.0 / 30 (86.7) & \textbf{29.0 / 30 (96.7)} & 28.0 / 30 (93.3) \\
Laplace-integral 1 & 25.0 / 30 (83.3) & \textbf{29.0 / 30 (96.7)} & 22.0 / 30 (73.3) \\
ODE & 7.0 / 54 (13.0) & 11.0 / 54 (20.4) & \textbf{14.0 / 54 (25.9)} \\
Polynomial roots & 12.0 / 63 (19.0) & \textbf{22.0 / 63 (34.9)} & 15.0 / 63 (23.8) \\
Nondim-symbolic & \textbf{65.0 / 65 (100.0)} & \textbf{65.0 / 65 (100.0)} & \textbf{65.0 / 65 (100.0)} \\
Nondim-numeric & 59.0 / 60 (98.3) & \textbf{60.0 / 60 (100.0)} & 59.0 / 60 (98.3) \\
\midrule
Overall & 221.0 / 362 (61.0) & \textbf{243.5 / 362 (67.3)} & 231.0 / 362 (63.8) \\
\bottomrule
\end{tabular*}
\end{table}

The HARDMath results are more mixed than PHYSICS, which is consistent with the benchmark's stricter symbolic answer format and the need for exact final expressions. Even after regrading for mathematical equivalence and partial credit, prompt remains strongest on several categories, while memory banks help most on ODE and tie the prompt baseline on one of the integral splits.

\subsection{GSM8K and MMLU capability checks}
\label{app:capability}

\begin{table}[H]
\centering
\small
\setlength{\tabcolsep}{5pt}
\caption{\textbf{Qwen3 GSM8K and MMLU capability checks.} These appendix results are intended as coarse capability-preservation checks rather than main steering results.}
\label{tab:gsm8k_mmlu}
\begin{tabular}{lllcccc}
\toprule
Benchmark & Model & Trait/Bank & Plain & Prompt & CAA & Ours \\
\midrule
GSM8K & Qwen3-30B-A3B & Warm & 0.970 & 0.960 & 0.970 & 0.960 \\
GSM8K & Qwen3-30B-A3B & Formal & 0.970 & 0.960 & 0.970 & 0.955 \\
GSM8K & Qwen3-30B-A3B & Direct & 0.970 & 0.950 & 0.975 & 0.950 \\
GSM8K & Qwen3-30B-A3B & Careful & 0.970 & 0.945 & 0.955 & 0.945 \\
GSM8K & Qwen3-30B-A3B & Curt & 0.970 & 0.910 & 0.955 & 0.955 \\
GSM8K & Qwen3-30B-A3B & Anxious & 0.970 & 0.935 & 0.965 & 0.940 \\
\midrule
MMLU & Qwen3-30B-A3B & Warm & 0.768 & 0.772 & 0.772 & 0.772 \\
MMLU & Qwen3-30B-A3B & Formal & 0.768 & 0.718 & 0.768 & 0.768 \\
MMLU & Qwen3-30B-A3B & Direct & 0.768 & 0.784 & 0.768 & 0.764 \\
MMLU & Qwen3-30B-A3B & Careful & 0.768 & 0.756 & 0.768 & 0.768 \\
MMLU & Qwen3-30B-A3B & Curt & 0.768 & 0.776 & 0.768 & 0.764 \\
MMLU & Qwen3-30B-A3B & Anxious & 0.768 & 0.706 & 0.768 & 0.764 \\
\bottomrule
\end{tabular}
\end{table}

Results on GSM8K and MMLU suggest that MI preserves capabilities, staying close to the plain-model performance across a range of traits and within the range already induced by Prompt and CAA.

\subsection{gpt-oss-120b personality results}
\label{app:gptoss}

The additional backbone and its attention implementation are described in Appendix~\ref{app:gptoss_impl}.

\begin{table}[H]
\centering
\small
\setlength{\tabcolsep}{5pt}
\caption{\textbf{Additional gpt-oss-120b personality results.}
Scores use the supplemental evaluation's drift coefficient of
1.5 and are reported separately from the coefficient-1 main-text
DAS/GAS. The five-trait aggregate is primary. The
openness-excluded row is a subset analysis using the same bank.}
\label{tab:gptoss_steering_supplement}
\begin{tabular}{llrr}
\toprule
Track & Reported metric & Prompt & MI \\
\midrule
Self-evaluation, five traits & $\mathrm{DAS}_{1.5}$ & 0.340 & 0.252 \\
Self-evaluation, excluding openness & $\mathrm{DAS}_{1.5}$ & 0.676 & 0.610 \\
Generation & $\mathrm{GAS}_{1.5}$ & 0.528 & 0.370 \\
\bottomrule
\end{tabular}
\end{table}

Prompt leads on every aggregate in
Table~\ref{tab:gptoss_steering_supplement}. The plain model scores
4.6 on openness, leaving limited room for a positive shift on
that trait. Excluding openness raises both methods' aggregate
scores but does not change their order.

\subsection{Outstanding ablations and omissions}
\label{app:ablations}

Figure~\ref{fig:layer_ablation} reports a Qwen3 layer-count sweep with fixed bank content. A placement control comparing selected and alternative sites at the same layer and unit budget is not reported. Such a control would separate the effect of site selection from the number of controlled layers.

\clearpage

\section{Technical derivations}
\label{app:technical_derivations}

\subsection{Bank-mixture derivation}
\label{app:bank-proof}

Let $\mathcal{I}_b$ index the slots belonging to bank $b$. Concatenated attention with bank-specific slot bias gives
\[
\widetilde\alpha_{t,m}^{(b)}
=
\frac{\exp(s_{t,m}^{(b)}+c_t^{(b)})}
{\sum_{b'}\sum_{r\in\mathcal{I}_{b'}}\exp(s_{t,r}^{(b')}+c_t^{(b')})}.
\]
Factor the numerator as
\[
\exp(s_{t,m}^{(b)}+c_t^{(b)})
=
\left(\sum_{r\in\mathcal{I}_b}\exp(s_{t,r}^{(b)})\right)
\frac{\exp(s_{t,m}^{(b)})}{\sum_{r\in\mathcal{I}_b}\exp(s_{t,r}^{(b)})}
\exp(c_t^{(b)}).
\]
Define the within-bank softmax
\[
\alpha_{t,m}^{(b)}
=
\frac{\exp(s_{t,m}^{(b)})}{\sum_{r\in\mathcal{I}_b}\exp(s_{t,r}^{(b)})}
\]
and bank evidence
\[
\widetilde\beta_t^{(b)}
=
\log\sum_{r\in\mathcal{I}_b}\exp(s_{t,r}^{(b)})+c_t^{(b)}.
\]
Then
$\widetilde\alpha_{t,m}^{(b)}
=\widetilde\pi_t^{(b)}\alpha_{t,m}^{(b)}$,
where
\[
\widetilde\pi_t^{(b)}
=
\frac{\exp(\widetilde\beta_t^{(b)})}{\sum_{b'}\exp(\widetilde\beta_t^{(b')})}.
\]
Therefore
\[
o_t^{\mathrm{concat},c}
=
\sum_b\sum_{m\in\mathcal{I}_b}\widetilde\alpha_{t,m}^{(b)}v_m^{(b)}
=
\sum_b\widetilde\pi_t^{(b)}\sum_{m\in\mathcal{I}_b}\alpha_{t,m}^{(b)}v_m^{(b)}
=
\sum_b\widetilde\pi_t^{(b)}u_t^{(b)}.
\]
This is the exact identity in Eq.~\eqref{eq:bank-mixture}.
Setting all $c_t^{(b)}$ to zero recovers ordinary concatenated
attention.

The size-normalized rule instead uses
$\beta_t^{(b)}=\widetilde\beta_t^{(b)}-\log M_b$ and
$\pi_t=\softmax([\beta_t^{(0)},\ldots,\beta_t^{(B)}])$.
Its slot weights are
\[
\alpha_{t,m}^{\mathrm{MI},(b)}
=
\frac{\exp(s_{t,m}^{(b)}+c_t^{(b)}-\log M_b)}
{\sum_{b'}\sum_{r\in\mathcal{I}_{b'}}
 \exp(s_{t,r}^{(b')}+c_t^{(b')}-\log M_{b'})}
=
\pi_t^{(b)}\alpha_{t,m}^{(b)}.
\]
The resulting output is
$o_t^\star=\sum_b\pi_t^{(b)}u_t^{(b)}$, as defined in
Eq.~\eqref{eq:normalized-bank-mixture}. Subtracting $\log M_b$
changes the bank probabilities when bank sizes differ.
It is equivalent to applying slot bias
$c_t^{(b)}-\log M_b$ in concatenated attention.

\subsection{\RoPE{} derivation}
\label{app:rope-proof}

With \RoPE{}, a prompt score between query position $t$ and key position $j$ is
\[
\frac{\langle R_t\bar q_t,R_j\bar k_j\rangle}{\sqrt{d_h}}
=
\frac{\langle \bar q_t,R_{j-t}\bar k_j\rangle}{\sqrt{d_h}}.
\]
For a canonical key $\tilde k_m$, assigning virtual relative phase $\delta_m$ gives
\[
\frac{\langle R_t^{-1}q_t,R_{\delta_m}\tilde k_m\rangle}{\sqrt{d_h}}
=
\frac{\langle \bar q_t,R_{\delta_m}\tilde k_m\rangle}{\sqrt{d_h}}.
\]
The inverse rotation cancels the query's explicit absolute rotary phase, with $\delta_m=0$ as the default relative phase. The canonical query, contextualized bank states, and generated behavior remain context-dependent.

\clearpage
\flushbottom
\section*{NeurIPS Paper Checklist}

\begin{enumerate}

\item {\bf Claims}
    \item[] Question: Do the main claims made in the abstract and introduction accurately reflect the paper's contributions and scope?
    \item[] Answer: \answerYes{}
    \item[] Justification: The abstract and introduction present memory inception as a training-free latent steering interface whose strongest evidence is the control--drift trade-off, Qwen3 dialogue updates, reasoning-bank transfer on PHYSICS and HARDMath, and selective KV savings. These claims match the benchmark protocols in Appendix~\ref{app:benchmarks} and the detailed results in Appendix~\ref{app:results_breakdown}.

\item {\bf Limitations}
    \item[] Question: Does the paper discuss the limitations of the work performed by the authors?
    \item[] Answer: \answerYes{}
    \item[] Justification: The conclusion includes dedicated limitations and responsible-release paragraphs covering benchmark diversity, bank quality, selector supervision, backbone dependence, and the gap between KV accounting and realized systems metrics.

\item {\bf Theory assumptions and proofs}
    \item[] Question: For each theoretical result, does the paper provide the full set of assumptions and a complete (and correct) proof?
    \item[] Answer: \answerYes{}
    \item[] Justification: The theory content is limited to method derivations, and Appendix~\ref{app:technical_derivations} includes the bank-mixture and \RoPE{} derivations together with the assumptions needed for those formulations.

    \item {\bf Experimental result reproducibility}
    \item[] Question: Does the paper fully disclose all the information needed to reproduce the main experimental results of the paper to the extent that it affects the main claims and/or conclusions of the paper (regardless of whether the code and data are provided or not)?
    \item[] Answer: \answerYes{}
    \item[] Justification: Section~\ref{sec:method} and Appendices~\ref{app:model_cards}--\ref{app:results_breakdown} specify the backbones, selector procedure, vLLM path, bank-construction rules, benchmark protocols, sample counts, and evaluation rubrics, and the abstract links to an anonymous repository with scripts and released benchmark assets.

\item {\bf Open access to data and code}
    \item[] Question: Does the paper provide open access to the data and code, with sufficient instructions to faithfully reproduce the main experimental results, as described in supplemental material?
    \item[] Answer: \answerYes{}
    \item[] Justification: The abstract provides an anonymous repository link for the released code, scripts, and benchmark assets used in the evaluation, together with accompanying repository documentation for reproducing the main experiments.

\item {\bf Experimental setting/details}
    \item[] Question: Does the paper specify all the training and test details (e.g., data splits, hyperparameters, how they were chosen, type of optimizer) necessary to understand the results?
    \item[] Answer: \answerYes{}
    \item[] Justification: The paper is inference-only and specifies the model backbones, steering baselines, benchmark construction, held-out evaluation rules, judge/grading protocols, selector freezing procedure, and benchmark sample counts used for the reported evaluations, especially in Appendix~\ref{app:benchmarks}.

\item {\bf Experiment statistical significance}
    \item[] Question: Does the paper report error bars suitably and correctly defined or other appropriate information about the statistical significance of the experiments?
    \item[] Answer: \answerNo{}
    \item[] Justification: The paper reports repeated grading and averaged judge scores for PHYSICS, along with full per-trait and per-task breakdowns elsewhere, but it does not provide confidence intervals, error bars, or formal significance tests for the main comparisons.

\item {\bf Experiments compute resources}
    \item[] Question: For each experiment, does the paper provide sufficient information on the computer resources (type of compute workers, memory, time of execution) needed to reproduce the experiments?
    \item[] Answer: \answerYes{}
    \item[] Justification: Appendix~\ref{app:model_cards} reports the inference stack and the GPU fleet used in practice (H100, H200, B200, and RTX Pro 6000 Blackwell). 
    
\item {\bf Code of ethics}
    \item[] Question: Does the research conducted in the paper conform, in every respect, with the NeurIPS Code of Ethics \url{https://neurips.cc/public/EthicsGuidelines}?
    \item[] Answer: \answerYes{}
    \item[] Justification: The work studies inference-time steering on existing models and does not involve human-subject data collection, deceptive annotation procedures, or undisclosed intervention on deployed users.

\item {\bf Broader impacts}
    \item[] Question: Does the paper discuss both potential positive societal impacts and negative societal impacts of the work performed?
    \item[] Answer: \answerYes{}
    \item[] Justification: The broader-impacts paragraph discusses beneficial uses such as persistent preferences and reusable heuristics, as well as misuse risks such as covert behavior shaping and hidden policy injection.
    
\item {\bf Safeguards}
    \item[] Question: Does the paper describe safeguards that have been put in place for responsible release of data or models that have a high risk for misuse (e.g., pre-trained language models, image generators, or scraped datasets)?
    \item[] Answer: \answerYes{}
    \item[] Justification: The conclusion's responsible-release paragraph states that the release is limited to anonymous code, evaluation scripts, and benchmark assets rather than a deployment-ready hidden-steering service, and it recommends disclosure, access control, audit logging, and human review for sensitive steering artifacts.

\item {\bf Licenses for existing assets}
    \item[] Question: Are the creators or original owners of assets (e.g., code, data, models), used in the paper, properly credited and are the license and terms of use explicitly mentioned and properly respected?
    \item[] Answer: \answerYes{}
    \item[] Justification: The paper credits the source models and benchmark releases, names the Llama 3.1 Community License and Apache 2.0 for the two backbones, notes that the IPIP questionnaire items are public-domain, and identifies provider service terms for API-based judges. 

\item {\bf New assets}
    \item[] Question: Are new assets introduced in the paper well documented and is the documentation provided alongside the assets?
    \item[] Answer: \answerYes{}
    \item[] Justification: The paper documents the dialogue benchmark construction, representative bank content, and evaluation setup in the appendix, and the linked anonymous repository provides the released implementation and usage documentation alongside these assets.

\item {\bf Crowdsourcing and research with human subjects}
    \item[] Question: For crowdsourcing experiments and research with human subjects, does the paper include the full text of instructions given to participants and screenshots, if applicable, as well as details about compensation (if any)? 
    \item[] Answer: \answerNA{}
    \item[] Justification: The paper does not use crowdsourcing or compensated human-subject experimentation.

\item {\bf Institutional review board (IRB) approvals or equivalent for research with human subjects}
    \item[] Question: Does the paper describe potential risks incurred by study participants, whether such risks were disclosed to the subjects, and whether Institutional Review Board (IRB) approvals (or an equivalent approval/review based on the requirements of your country or institution) were obtained?
    \item[] Answer: \answerNA{}
    \item[] Justification: No human-subject study requiring IRB review was conducted.

\item {\bf Declaration of LLM usage}
    \item[] Question: Does the paper describe the usage of LLMs if it is an important, original, or non-standard component of the core methods in this research? Note that if the LLM is used only for writing, editing, or formatting purposes and does \emph{not} impact the core methodology, scientific rigor, or originality of the research, declaration is not required.
    \item[] Answer: \answerYes{}
    \item[] Justification: The evaluation methodology explicitly uses external LLM judges: GPT-4o-mini for Big Five generation and dialogue/long-context judging, and GPT-5.5-thinking for HARDMath and PHYSICS grading, as described in Appendix~\ref{app:benchmarks}.

\end{enumerate}

\end{document}